\documentclass[journal]{IEEEtran}
\usepackage{cite}
\usepackage{amsmath,amssymb,amsfonts}
\usepackage{epstopdf}
\usepackage{graphicx}
\usepackage{textcomp}
\usepackage{hyperref}
\usepackage{floatrow}
\usepackage[ruled]{algorithm2e}
\usepackage{algpseudocode}
\usepackage{float}
\usepackage{multirow} 
\usepackage{subfigure}
\usepackage{threeparttable}
\usepackage{longtable,lscape}
\usepackage[table,xcdraw]{xcolor}
\usepackage{footnote}
\usepackage{booktabs}
\newfloatcommand{capbtabbox}{table}[][\FBwidth]
\usepackage{url}

\begin{document}

\title{Graph Neural Network Encoding for Community Detection in Attribute Networks}

\author{Jianyong~Sun~\IEEEmembership{Senior Member,~IEEE}, Wei~Zheng, Qingfu Zhang~\IEEEmembership{Fellow, IEEE} and Zongben Xu
\thanks{J. Sun, W. Zheng, and Z. Xu are with the School of Mathematics and Statistics, Xi'an Jiaotong University, Xi'an 710049, China.
e-mail: jy.sun@xjtu.edu.cn, weizheng@stu.xjtu.edu.cn and zbxu@xjtu.edu.cn.}
\thanks{Q. Zhang is with Department of Computer Science, The City University of Hong Kong, Hong Kong, China. email: qingfu.zhang@cityu.edu.hk.}
}

\markboth{}%
{Sun, Zheng, Zhang and Xu: Graph Neural Network Encoding for Community Detection in Attribute Networks}

\maketitle

\begin{abstract}
In this paper, we first propose a graph neural network encoding method for multiobjective evolutionary algorithm to handle the community detection problem in complex attribute networks. In the graph neural network encoding method, each edge in an attribute network is associated with a continuous variable. Through non-linear transformation, a continuous valued vector (i.e. a concatenation of the continuous variables associated with the edges) is transferred to a discrete valued community grouping solution. Further, two objective functions for single- and multi-attribute network are proposed to evaluate the attribute homogeneity of the nodes in communities, respectively. Based on the new encoding method and the two objectives, a multiobjective evolutionary algorithm (MOEA) based upon NSGA-II, termed as continuous encoding MOEA, is developed for the transformed community detection problem with continuous decision variables. Experimental results on  single- and multi-attribute networks with different types show that the developed algorithm performs significantly better than some well-known evolutionary and non-evolutionary based algorithms. The fitness landscape analysis verifies that the transformed community detection problems have smoother landscapes than those of the original problems, which justifies the effectiveness of the proposed graph neural network encoding method.
\end{abstract}


\IEEEpeerreviewmaketitle

\section{Introduction}\label{a}

\IEEEPARstart{A} graph network can be represented as a set of nodes (vertices) and edges that connect these nodes. Complex networks have been used to model many real-world network systems, such as the World Wide Web~\cite{www}, scientific collaboration networks~\cite{NewmanPNAS2001}, social and biological networks~\cite{GirvanPNAS2002}, and many others, since these networks all exhibit some community structures. Unveiling these structures, also called community detection, is thus of great importance to understand the behavior and organization of complex networks, and the relationships among generic entities.

The goal of community detection is to partition all nodes in a complex network into some clusters such that nodes within a cluster are densely connected to each other and sparsely to nodes in other clusters. This problem has been proved to be NP-hard~\cite{PhysRevE76}. Due to the importance of the complex network detection problem, research on this subject has become popular since 1930s~\cite{Fortunato2010}. A large interdisciplinary community of scientists have been working on this problem and a large amount of methods have been proposed for different types of complex networks. Surveys of community detection in graphs and networks can be found in every several years from 2005 until recently~\cite{Communityreview2018,attibutedcommunitysurvey}. In this paper, we do not intend to review all literatures but only on approaches based on evolutionary algorithm (EA) which is closely related to our work.

Various metrics have been proposed to quantitatively measure the quality of a partition to a graph network~\cite{Communityreview2018}, e.g. the modularity ($Q$)~\cite{Q}, the community score ($CS$)~\cite{GA-Net}, the sum of community fitness $\cal{P(S)}$~\cite{P-S}, and others. The community detection problem can then be formalized as a discrete optimization problem based on the optimization of one or several metrics. As a promising paradigm for discrete optimization, EAs or multi-objective EAs (MOEAs) have also been applied for this problem, please refer to a recent survey in~\cite{2018reviewEA}.

Genetic algorithm was firstly adopted in~\cite{first2007community} for maximizing $Q$. Since then, several genetic algorithms (GA), including MIGA~\cite{MIGA}, MAGA-Net~\cite{MAGANet} and Meme-Net~\cite{MemeNet}, and GDPSO~\cite{GDPSO}, were also developed based on optimizing $Q$, while GA-Net~\cite{GA-Net} was proposed based on optimizing $CS$.

MOEAs have also been applied since it is not comprehensive to measure a partition solution by only a single objective. The first multiobjective genetic algorithm, dubbed as MOGA-Net~\cite{conMOGA-Net}\cite{MOGA-Net}, was proposed in 2009. It was built upon NSGA-II in which two objectives including $CS$ and $\cal{P(S)}$ are used. MOEA/D-Net~\cite{MOEA/D-Net} takes the Negative Ratio Association (NRA) and Ratio Cut (RC) as two objectives, which is built upon the framework of multi-objective evolutionary algorithm based on decomposition (MOEA/D). MICD~\cite{MICD}, MODBSA~\cite{MODBSAD}, and DIM-MOEAD~\cite{DIMMOEAD}, were all developed based on the two objectives, but under different MOEA frameworks. In MOCD~\cite{MOCD} and MMCD~\cite{MMCD}, two objectives obtained by decomposing the modularity $Q$, which are to measure the intra-cluster edge density and inter-cluster sparsity, were used. Besides these works, MOEAs have also been applied on large networks~\cite{CYB2020largenework}, signed networks~\cite{CYB2014MOEAsignednetwork} and dynamic networks~\cite{CYB2020Dynamic}, in which various metrics for specific networks are developed and optimized.

In this paper, we focus on the community detection for complex attribute networks. In many real complex networks, besides the connecting edges among nodes, there are also attributes associated with each node which are to describe the node's properties. For example, in a social network, each user may have attributes like age, sex, degree, hobby, and other tags. Such networks are often called attributed complex networks~\cite{attibutedcommunitysurvey}. For such a network, community detection requires to reveal not only the distinct network topological structure, but the homogeneity of attributes within clusters. For example, we may wish to find a group of users with similar hobbies. The extra homogeneity requirement makes the community detection problem for attribute network much more difficult.

For attribute complex networks, revealing network structure and node attributes' homogeneity are two desired goals. Through establishing appropriate objectives for network structure and node attributes, MOEAs could also be applied for the attribute complex network community detection problems.

To the best of our knowledge, only two promising papers based on MOEAs have been published for attribute network detection. The first one is MOEA-SA~\cite{MOEA-SA}, in which a new objective $S_A$ was proposed to measure the attribute similarity within clusters. Together with the modularity~\cite{Q}, MOEA-SA is developed upon NSGA-II~\cite{NSGA-II} with a hybrid network encoding method and a multi-individual-based mutation operator. Besides, a neighborhood correction strategy is proposed to repair improper solution. The other one is MOGA-@Net~\cite{MOGA-@Net}, which is also developed based on NSGA-II. In MOGA-@Net, three objectives (namely, modularity~\cite{Q}, community score~\cite{GA-Net} and conductance~\cite{conductance}) for evaluating the structural dimension and three objectives (namely, Jaccard, cosine and Euclidean-based similarity) for measuring the attribute homogeneity are considered. A post-processing local merge procedure is further introduced to merge the communities.

In the above MOEA-based community detection algorithms, an encoding process is required to initialize individuals and a decoding process to retrieve individuals to their corresponding partitions for evaluating their qualities. There are two widely used encodings, including the locus-based~\cite{locusbased} and label-based~\cite{first2007community}. It is argued in~\cite{MOEA-SA} that the locus-based encoding is able to initialize for good individuals, but is time-consuming when decoding. The label-based encoding, on the other hand, is easy for designing evolutionary operators, but is not good at initialization since the adjacency information is not involved.

In this paper, we propose a novel encoding method. The new encoding method is implemented by first associating each edge in a graph network with a continuous variable, then transforming the concatenation of the continuous variables to a partition solution of the considered attribute network by a series of non-linear functions. Based on this encoding, the original discrete-valued community detection problem is transformed into a continuous one. We then propose a continuous-coded MOEA built upon NSGA-II~\cite{NSGA-II}, in which each individual is a continuous valued vector as opposed to a discrete valued vector in the locus-based and label-based encodings.

There are mainly two reasons that motive us to develop such transformation. First, as different to continuous problems, there is no sufficient and useful neighborhood information to help searching in discrete problems~\cite{Maehara2018}. Second, the local structure of a discrete problem is usually with a high ruggedness which means that the fitness landscape is not smooth. That is, a small change of the genotype may result in a substantial change of the phenotype~\cite{Peng14,Rothlauf06}. As a result, this may cause oscillation of the search process~\cite{Chatzarakis18}. These factors make it very difficult to search over the discrete search space. On the contrary, the proposed continuous encoding method can make the fitness landscape of the transformed problem smoother than the original landscape. This will not only make the search easier, but also remedy the shortcomings of the locus and label-based encoding methods. To verify these grounds, a fitness landscape analysis is carried out in this paper. The analysis confirms that the transformed continuous problem indeed has a smoother landscape than that of the original community detection problem.

Further, to better measure the attribute homogeneity in the communities, we propose two objectives, similar to those proposed in~\cite{MOEA-SA}, to handle the single- and multi-attribute similarity, respectively.

The rest of this paper is organized as follows. Section~\ref{r} introduces some preliminaries, including the definition of attribute complex network, relevant concepts in multiobjective optimization and MOEA. The proposed method, including the graph neural network encoding, the objectives for attributes and the continuous encoding MOEA (CE-MOEA), is presented in Section~\ref{c}. Experiment studies on a variety of networks with different types are carried out in Section~\ref{d}. The fitness landscape analysis is presented in Section~\ref{g}. Related work on non-EA based community detection is reviewed in Section~\ref{rw}. Section~\ref{h} concludes this paper.

\section{Preliminaries}\label{r}

\subsection{Complex Attribute Network}\label{b.1}

An attribute network is a 3-tuple ${\cal G} = ({\cal V}, {\cal E}, {\cal A})$, where ${\cal V} = \{V_1, V_2, \cdots, V_r\}$ is the set of nodes, ${\cal E} = \{e_{ij}: 1\leq i ,j \leq r\}$ is the set of edges ($e_{ij} = 1$ means $V_i$ links to $V_j$), ${\cal A} = \{a_1, a_2, \cdots, a_r\}$ is the set of attributes for the nodes. Here $a_i, 1\leq i \leq r$ can be discrete or continuous, and may be one or multiple dimensional.

\begin{figure}[htbp]
 \includegraphics[scale=0.21]{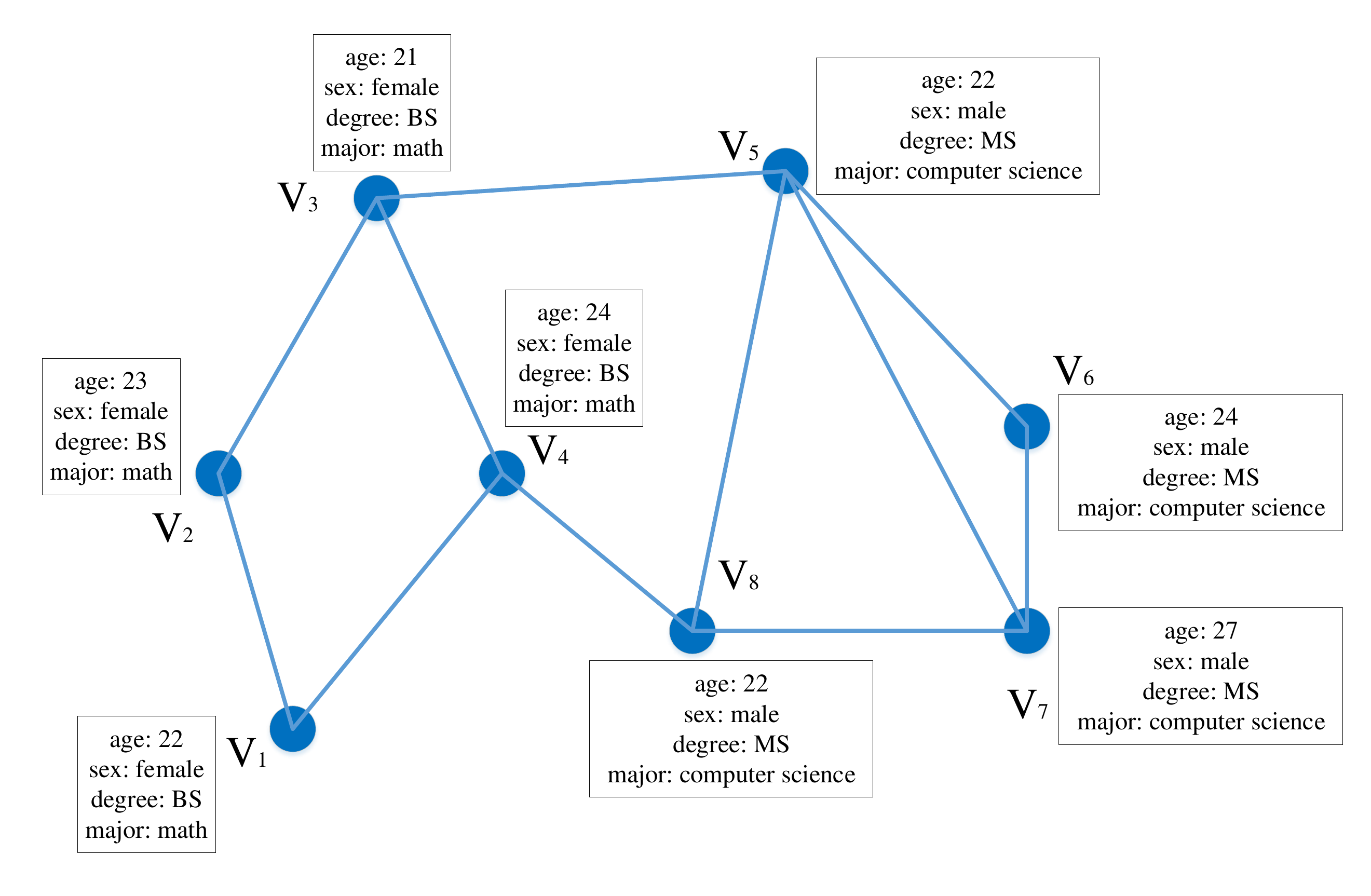}
\caption{An example attribute network with 8 nodes, 10 edges. There are 4 attributes for each node.}\label{fig1}
\end{figure}

Fig.~\ref{fig1} shows a simple attribute network example. The network has 8 nodes and 10 edges. Each node has 4 attributes (age, sex, degree, and major). According to the attributes of each node, it is seen that this network can be divided into two communities: $\{V_1, V_2, V_3, V_4\}$ and $\{V_5, V_6, V_7, V_8\}$. However, if consider only the ``age" attribute, it is rather difficult to partition this network.

On the other hand, it is also not easy to partition the network based purely on its structure. However, considering both attributes and network structure, it might be easy to determine two communities: $\{V_1, V_2, V_3, V_4\}$ and $\{V_5, V_6, V_7, V_8\}$. This partition not only minimizes the similarity within communities, but maximizes the communities' attributes homogeneity.

\subsection{Multiobjective Optimization}\label{b.2}

A multi-objective optimization problem (MOP) can be stated as follows:
\begin{equation}\label{1}
\begin{split}
&\textrm{minimize\ } F(\mathbf{w})=(f_1(\mathbf{w}),f_2(\mathbf{w}),...,f_m(\mathbf{w}))^\intercal\\
&\textrm{subject to\ } \mathbf{w}\in\boldsymbol\Omega
\end{split}
\end{equation}where $\boldsymbol\Omega$ is the search space (could be continuous or discrete), $\mathbf{w} = (w_1,\ldots,w_n)\in \boldsymbol\Omega$ is the decision variable. $F:\boldsymbol\Omega\rightarrow \mathbb{R}^m$ consists of $m$ real-valued objective functions.

In the MOP taxonomy, a vector $\mathbf{x} = (x_1, \cdots, x_m)$ is said to dominate a vector $\mathbf{y} = (y_1, \cdots, y_m)$ denoted as $\mathbf{x} \prec \mathbf{y}$) if and only if there exists at least one $k$ such that $x_j \leq y_j$, $\forall j \in \{1, \cdots, m\}$ but $x_k < y_k$. If a solution $\mathbf{x^\ast} \in \boldsymbol\Omega$ is not dominated by any other solution, $\mathbf{x^\ast}$ is called a Pareto optimal solution. There exists many optimal solutions that are non-dominated to each other. The set of all these optimal solutions is called the Pareto Set (PS), while its image is called the Pareto Front (PF).

The primal advantage of the MOEA paradigm is that an approximated PS can be reached in a single run. The study of MOEA is one of the most popular avenues in computational intelligence. There are main four categories of MOEAs, namely Pareto dominance relation based (such as NSGA-II~\cite{NSGA-II} and NSGA-II/CSDR~\cite{NSGA-IICSDR}), performance metric based (such as HypE~\cite{Hype} and FV-MOEA~\cite{FVMOEA}), decomposition based (such as MOEA/D~\cite{MOEAD}, MOEA/D-IR~\cite{MOEADIR} and MOEA/D-CMA~\cite{MOEADCMA}) and learning based MOEAs (such as OCEA~\cite{Jianyong18}, CA-MOEA~\cite{CAMOEA} and GMOEA~\cite{GMOEA}). We do not intend to review the rich literature of MOEA in this paper. Interested readers please refer to~\cite{MOEAsurvey}.

In this paper, the purpose of attribute complex network detection problem is to find a partition of communities such that two requirements are satisfied, including 1) the edges between communities are sparse and those within the community are dense; and 2) the node attributes in the same community should be similar as much as possible while the similarity of node attributes in different communities should be dissimilar. Therefore, the community detection problem can be readily modeled as a two-objective optimization problem. Using MOEA to solve this problem is thus straightforward and maybe promising.

\section{The Method}\label{c}

In this section, the graph neural network encoding method is first presented, followed by two newly developed objectives for attribute homogeneity, and CE-MOEA.

\subsection{Graph Neural Network Encoding}\label{c.1}

\begin{figure}
 {\includegraphics[scale=0.3]{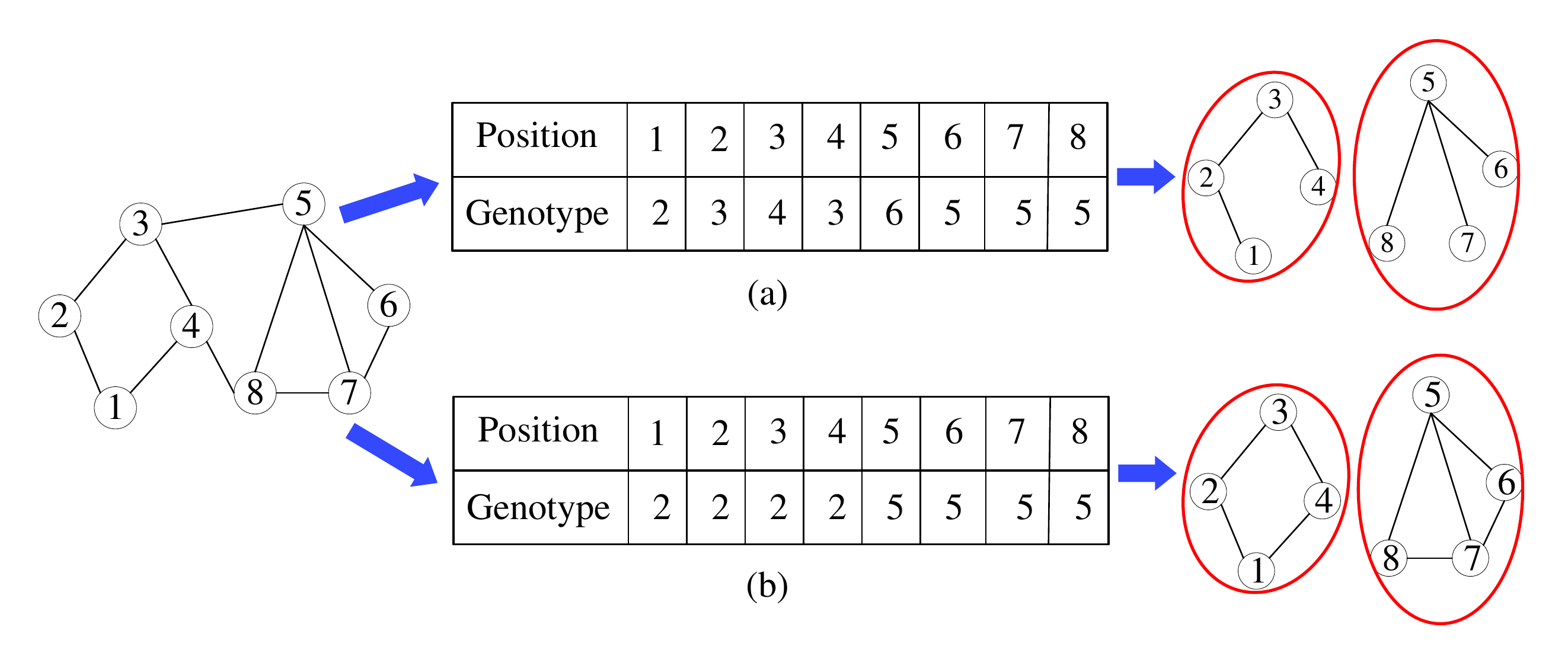}}
\caption{An example of the locus-based encoding and label-based encodings. (a) Locus-based encoding. (b) Label-based encoding.}\label{fig2}
\end{figure}

The locus-based~\cite{locusbased} and label-based~\cite{first2007community} encodings have been widely used in MOEAs for network related optimization problem. Fig.~\ref{fig2} shows an example of the two encodings for the network in Fig.~\ref{fig1}. It is seen that both encodings have a coding length equivalent to the number of nodes in the network.

In the locus-based encoding, a node's genotype is taken as one of its linked nodes. For example, in the example network, node 1 links to node 2 and 4. The genotype of node 1 could thus be 2 or 4. The shown individual genotype $(2,3,4,3,6,5,5,5)$ in Fig.~\ref{fig2}(a) is obtained by associating each node with one of its linked nodes. This individual can be decoded into two communities, i.e. $\{1,2,3,4\}$ and $\{5,6,7,8\}$, by simply retrieving it to an induced graph to the original one.

In the label-based encoding, each node's genotype can be any integer in $\{1,2,\cdots,r\}$. This integer indicates which cluster this node belongs to. As shown in Fig.~\ref{fig2}(b), 2 and 5 are selected as the genotype of the nodes. The decoding process is to simply take the nodes with the same cluster index together. It is seen that the same partition of communities as the previous encoding are obtained after decoding. It should be noted that the resultant genotypes by the two encoding methods are all discrete vector.

As argued in~\cite{MOEA-SA}, it is difficult to design evolutionary operators for the locus-based encoding, and difficult to initialize individuals with high quality for the label-based encoding since the adjacency information among nodes is not used.

In the following, we present the proposed graph neural network encoding method. We summarize its pseudo code in Alg.~\ref{alg1}. In Alg.~\ref{alg1}, a continuous valued vector $\mathbf{x} \in \mathbb{R}^{d}$ where $d = \sum_{i,j} e_{ij}$ is the number of edges, is taken as the input. $\mathbf{x}$ is a concatenation of $r$ sub-vectors, where $\mathbf{x}_i$ represents the continuous vector associated with node $V_i, 1\leq i\leq r$. The length of $\mathbf{x}_i$ is $d_i = \sum_j e_{ij}$. That is, each link connecting $V_i$ to the other nodes is assigned with one continuous value. We denote the set of nodes that links with $V_i$ as $D_i$.

For node $V_i$, denote $\mathbf{x}_i = [x_{i,1},\cdots, x_{i,d_i}]$, we first apply a sigmoid function $\sigma$ which is defined as follows:\begin{equation}
\sigma(x) = \frac{1}{1+\exp(-x)}
\end{equation}over $\mathbf{x}_i$ element by element. This gives $\mathbf{h}_i \in (0,1)^{d_i}$ (line~\ref{sigma}). A softmax function is then applied on $\mathbf{h}_i$ to obtain $\mathbf{p}_i = [\mathbf{p}_{i,1}, \cdots, \mathbf{p}_{i,d_i}]$ (line~\ref{softmax}) where
\begin{equation}
\mathbf{p}_{ij} = \frac{\exp(\mathbf{h}_{ij})}{\sum_j \exp(\mathbf{h}_{ij})}, 1\leq j \leq d_i.
\end{equation}
Since $\mathbf{p}_{ij}\geq 0$ and $\sum_j \mathbf{p}_{ij} = 1$, this actually gives the probability of choosing a node from $D_i$. We propose to choose node $s_i$ such that
\[s_i = \arg\max_{j = 1, \cdots, d_i} \mathbf{p}_{ij}\]i.e. the $\text{argmax}$ operation as seen in line~\ref{argmax}. This means that node $V_i$ is linked to $V_{s_i}$ in the genotype. The above process is carried out for all nodes in the considered network to obtain the set of nodes (i.e. $\mathcal{S}$ in line~\ref{retuns}) and the set of edges (i.e. $\mathcal{E}_{\mathcal{S}}$ in line~\ref{edgein}). With the obtained $\mathcal{S}$ and $\mathcal{E}_{\mathcal{S}}$, a partition ${\cal G}_{\mathcal{S}}$ can be returned after decoding (line~\ref{decoding}).
\begin{algorithm}[htbp]
\small{
\caption{Graph Neural Network Encoding Method}
\label{alg1}
\LinesNumbered
\KwIn{$\mathbf{x} = [\mathbf{x}_1, \cdots, \mathbf{x}_r] \in \mathbb{R}^{d}$.}
\KwOut{A community partition ${\cal G}_{\mathcal{S}}$.}
Set $\mathcal{S} = \emptyset$ and $\mathcal{E}_{\mathcal{S}} = \emptyset$;\ \\
\For{$i \leftarrow 1 \text{\ to\ } r$}{
$\mathbf{h}_i \leftarrow  \sigma(\mathbf{x}_i)$; \label{sigma} \\
$\mathbf{p}_i \leftarrow \text{softmax}(\mathbf{h}_i)$; \label{softmax} \\
$s_i\leftarrow \text{argmax}(\mathbf{p}_{i})$; \label{argmax}\\
$\mathcal{S} \leftarrow \mathcal{S} \bigcup s_i$;\label{retuns} \\
$\mathcal{E}_{\cal S} \leftarrow \mathcal{E}_{\mathcal{S}} \bigcup e_{V_i,V_{s_i}}$;\label{edgein} \\
}
\Return ${\cal G}_{\mathcal{S}} \leftarrow \text{Decoding}(\mathcal{S}, \mathcal{E}_{\mathcal{S}})$.\label{decoding}
}
\end{algorithm}

\begin{figure}
\includegraphics[scale = 0.25]{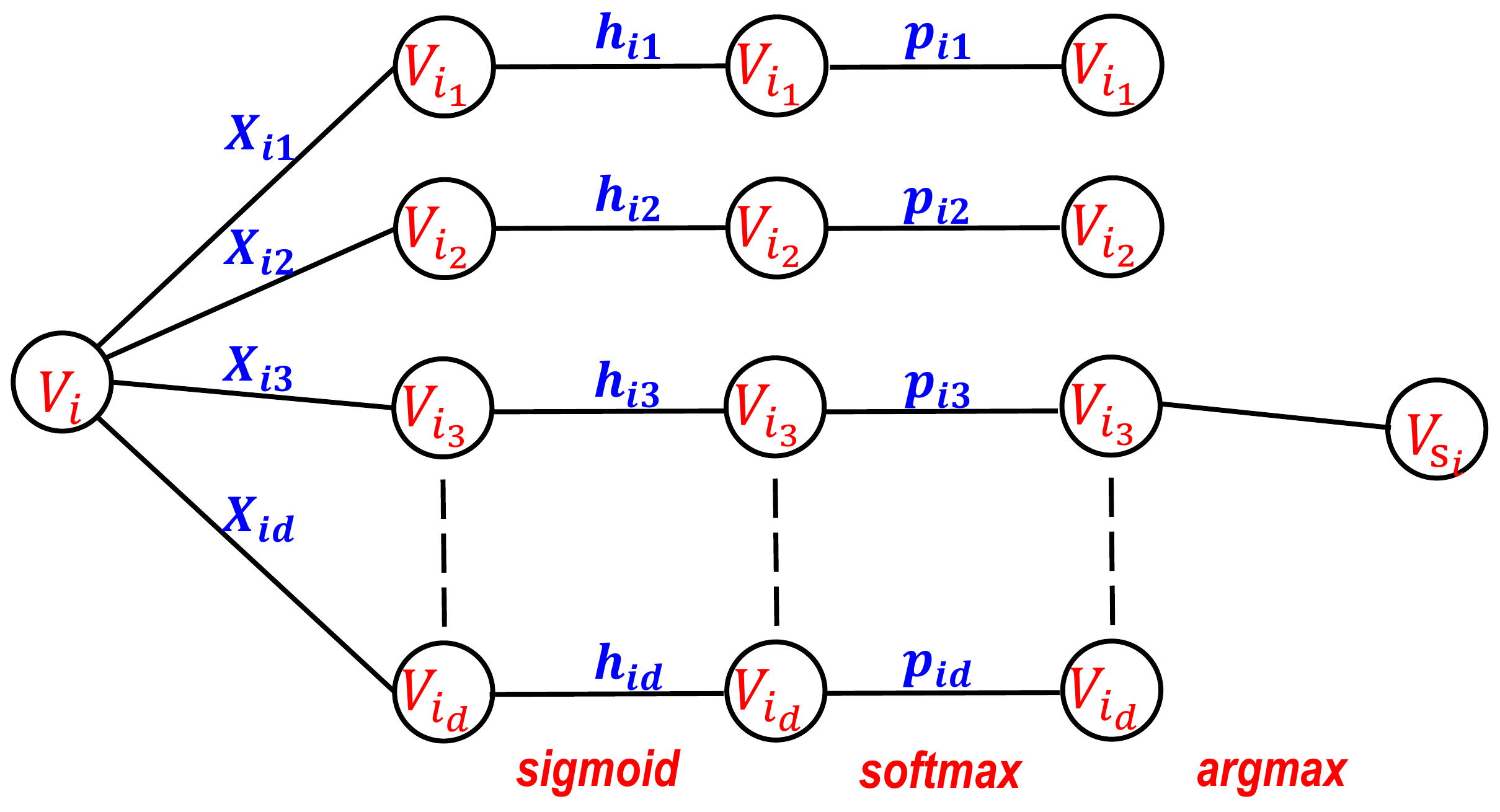}
\caption{Demo of the graph neural network encoding of a single node $V_i$.}\label{singlenode}
\end{figure}

Fig.~\ref{singlenode} shows the encoding process of a single node $V_i$. From the figure, it is seen that for each $V_i$, there associates a continuous value for each node in $D_i$. Through sigmoid, softmax and argmax operation, node $V_{s_i}$ is selected to be linked to $V_i$ in the genotype.

\begin{figure*}
\includegraphics[scale = 0.5]{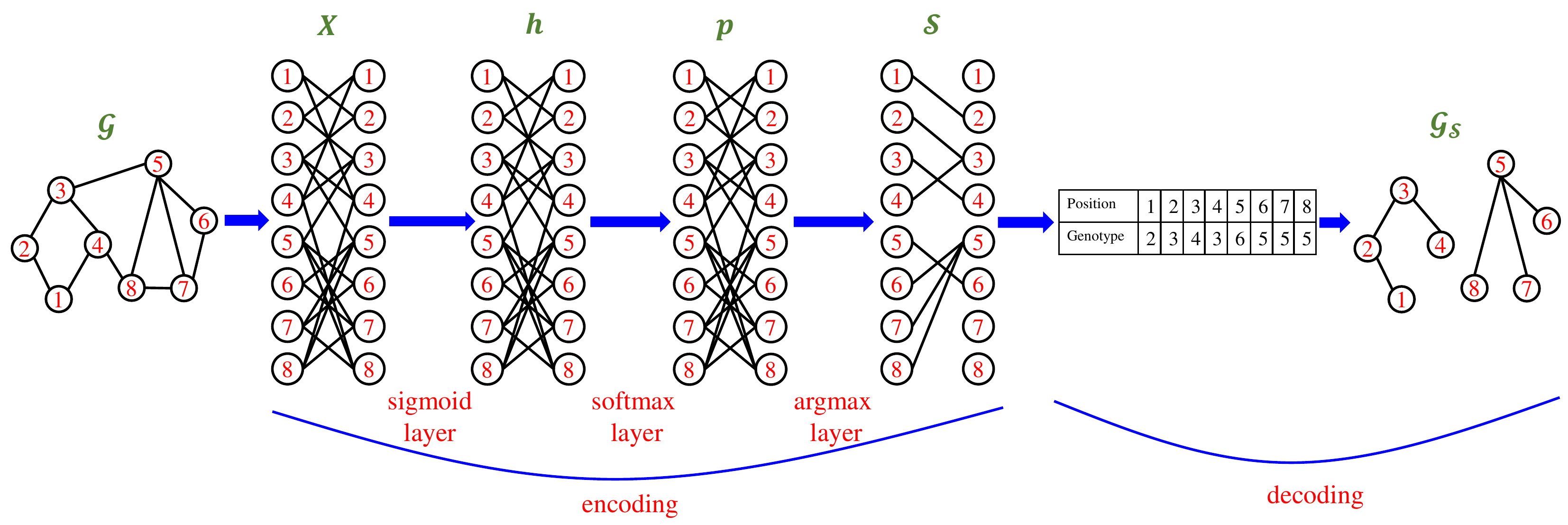}
\caption{Demo of the graph neural network encoding and decoding process.}\label{fullprocess}
\end{figure*}

Fig.~\ref{fullprocess} shows the full process of encoding and decoding taking the network in Fig.~\ref{fig1} as an example. Given the network (denoted as $\cal G$), with continuous vector $\mathbf{x}$ associated with the edges, the sigmoid operation (which can be regarded as the sigmoid layer in neural network) is applied to obtain $\mathbf{h}$. The softmax layer is then applied on $\mathbf{h}$ to obtain $\mathbf{p}$. Through argmax operation (layer), each node is linked to the node that is with the greatest probability entity in its corresponding $\mathbf{p}$ values. This concludes the encoding process, which lead to a locus-based representation. The decoding process can thus turn the representation into a partition ${\cal G}_{\mathcal{S}} $ of $\cal G$.

In the following, for the sake of simplicity, we use \begin{equation}\label{gne}{\cal G}_{\mathcal{S}} = \text{GNN}(\mathbf{x}; {\cal G})\end{equation} to represent the encoding of a genotype to a network $\cal G$. That is, given $\mathbf{x}$, a network partition ${\cal G}_{\mathcal{S}}$ can be obtained by function \text{GNN}$(\cdot)$. With the obtained network partition, objectives such as the modularity can be calculated.

\subsection{The Objective Functions}\label{c.2}

\subsubsection{Objective Regarding the Network Structure}

The well-known modularity $Q$ proposed in~\cite{Q} is used as the first objective in our study for revealing the network structure. Given a network ${\cal G}$ and its partition ${\cal G}_{\mathcal{S}}$, let $c$ be the number of obtained communities, $l_k$ be the total number of edges that connect the nodes within the community $k$, $d_k$ is the sum of degrees of nodes of community $k$ and $L$ stands for the total number of edges, the modularity $Q$ is defined as:
\begin{equation}
 Q = \sum_{k = 1}^{c}\bigg[\frac{l_k}{L} - \bigg(\frac{d_k}{2L}\bigg)^2\bigg]\triangleq f_Q ({\cal G}_{\mathcal{S}}; {\cal G})\label{2}
\end{equation} A higher $Q$ value indicates a network with a more well-defined community structure.

Together with Eq.~\ref{gne}, given $\mathbf{x}$, the modularity can be computed by composing functions $f_Q$ and $\text{GNN}$ as follows:
\[ Q =  f_{Q}\circ \text{GNN}(\mathbf{x}; {\cal G})\]For the sake of simplicity, we denote $Q(\mathbf{x})$ as the function to compute the modularity taking $\mathbf{x}$ as the decision variable.

\subsubsection{Objective Evaluating Attribute Similarity}

To measure the difference between two nodes' attributes, the following two objectives are used for single- and multi-attribute homogeneity, which is a modification to those objectives proposed in MOEA-SA~\cite{MOEA-SA}, respectively.

For a single-attribute network with real-valued attributes, a similarity objective function $f_s$ is proposed as follows:
\begin{equation}
f_s = \frac{S_O}{\sum_{k = 1}^{c} r_k(r_k-1)}\label{3}
\end{equation}
where
\begin{equation}
 \small S_O = \sum_{k = 1}^{c}\mathop{\sum_{V_i,V_j\in C_k}}\limits_{i < j}\sqrt{(a_i - a_j)^2}
\end{equation} where $c$ is the number of obtained clusters, $C_k$ is the cluster $k$, $r_k$ stands for the number of nodes within cluster $k$, $a_i$ (resp. $a_j)$ is the attribute of $V_i$ (resp. $V_j$). $S_O$ is the sum of the Euclidean distance between node attributes of each community. The denominator is the summation of all obtained clusters of values $r_k(r_k-1)$. Note that in MOEA-SA, the numerator is $\sum_{k = 1}^{c}\sum_{i,j\in k,i<j}2s(i,j)$, where $s(i,j) = 0$ while $a_i = a_j$ and $s(i,j) = 1$ while $a_i \neq a_j$. It is to measure the distance between single attribute homogeneity within the detected communities.

For a multi-attribute network with binary attribute values, a cosine-based similarity objective function $f_m$ is proposed to measure the attribute similarity:
\begin{equation}
f_m = \frac{M_O}{\sum_{k = 1}^{c} r_k(r_k-1)}\label{4}
\end{equation}where
\begin{equation}
\small M_O = \sum_{k = 1}^{c}\mathop{\sum_{V_i,V_j\in C_k}}\limits_{i<j}\frac{a_i\cdot a_j}{\|a_i\|\|a_j\|}
\end{equation}where $\|\cdot\|$ means the norm of a vector. The numerator is the cosine value of attributes of each node pair's attributes within a community $k$. The summation of all detected clusters is denoted as $M_O$.  The denominator is the same as in $f_s$.

It can be found that in $f_s$ (or $f_m$), the smaller the value of $f_s$ (or $f_m$) is, the more homogeneous of the node attributes in the obtained communities is. Therefore, the node attribute clustering problem can be viewed as a problem of finding a division of a network such that the attribute similarity objective function $f_s$ or $f_m$ is minimized. Similar to the definition of $Q(\mathbf{x})$, we also define $f_s(\mathbf{x})$ and $f_m(\mathbf{x})$.

In summary, based on the proposed graph neural network encoding, given a continuous valued vector $\mathbf{x}$, the modularity and attribute similarity can be computed. The problem is thus to find an approximation set to PS w.r.t. $\mathbf{x}$ such that the objective vector $F = (-Q(\mathbf{x}), f_s(\mathbf{x}))$ (or $F = (-Q(\mathbf{x}), f_m(\mathbf{x}))$) is minimized. Formally, the community detection problem for attribute network can be defined as follows:
\begin{eqnarray}\label{problem}
\begin{split}
&&\text{minimize\ } F = (-Q(\mathbf{x}), f_s(\mathbf{x})) \\
&&  \text{or\ } F = (-Q(\mathbf{x}), f_m(\mathbf{x})) \\
&& \text{\ s.t.\ } \mathbf{x} \in {[0,1]^{d}}
\end{split}
\end{eqnarray}
Here the reason to set $ \mathbf{x} \in {[0,1]^{d}}$ is to make the range of the sigmoid controllable and the softmax is scale-invariant.


\subsection{The Algorithm}\label{c.3}
\begin{figure}
\includegraphics[scale = 0.25]{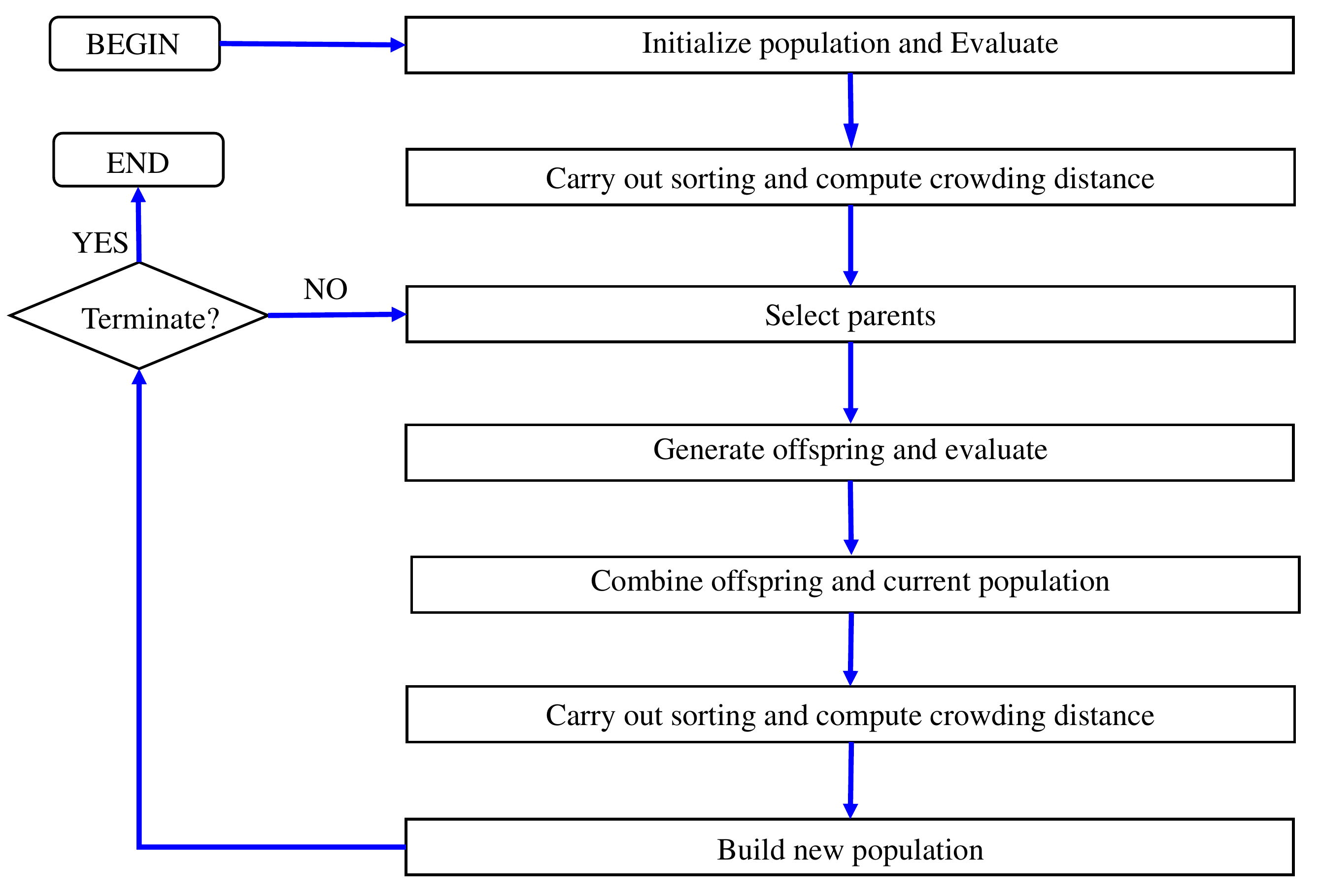}
\caption{The flow chart of CE-MOEA.}\label{flowchart}
\end{figure}

The developed algorithm is built upon the well-known Non-dominated Sorting Genetic Algorithm II (dubbed as NSGA-II) with differential evolution (DE) operators~\cite{DE}. Its flow chart is shown in Fig.~\ref{flowchart} and the pseudocode is summarized in Alg.~\ref{alg2}.

\begin{algorithm}[htbp]
\small{
\caption{The Non-dominated Sorting Method}
\label{alg4}
\LinesNumbered
\KwIn{Population $P$.}
\KwOut{$L_s$: some nondominated layers.}
Set index $i = 1$;\ \\
\While{$P \neq \emptyset$}{
Find all solutions $S$ who are not dominated by any solution in $P$;\ \\
$L_{s_i} \leftarrow S$ and $P \leftarrow P\setminus S$;\ \\
$i \leftarrow i + 1$;\ \\
}
}
\end{algorithm}

The functions $\text{FNS}(\cdot)$, $\text{CWD}(\cdot)$ and $\text{BTS}(\cdot)$ in Alg.~\ref{alg2} represent the fast non-dominated sorting, crowding distance, and binary tournament selection, respectively. They are just standard operations used in NSGA-II. The $\text{FNS}(\cdot)$ function sorts solutions into several non-dominated layers based on their dominance relationship. Its pseudocode is shown in Alg.~\ref{alg4} (taken from~\cite{NSGA-II}). The $\text{CWD}(\cdot)$ function is used to maintain the diversity of population. The $\text{DE}(\cdot)$ function stands for the differential evolution operation which will be described later. More details of NSGA-II please refer to~\cite{NSGA-II}.

In Alg.~\ref{alg2}, the first population $P_1$ is randomly initialized in $[0,1]^d$ (line~\ref{ce1}), and individuals  are evaluated according to Eq.~(\ref{problem}). Their objectives are organized in $\cal F$ (line~\ref{ce2}). The non-dominated layers and the crowding distances of ${\cal F}$ are then computed in line~\ref{ce3}. From line~\ref{ce4} to~\ref{ce16}, the NSGA-II operations are performed to optimize Problem~(\ref{problem}). The binary tournament selection is carried out on $P_g$ to obtain a parent set $\cal P$ based on $\cal F$ and the crowding distance $C_d$ (line~\ref{ce5}). By selecting parent individuals from $\cal P$, $\text{DE}$ is applied to generate new offsprings (line~\ref{ce9}). The newly generated individuals are evaluated in line~\ref{ce10} and combined with current population (line~\ref{ce11}). The combined individual objectives are sorted to obtain the non-dominated layers and the crowding distances (line~\ref{ce12}-\ref{ce13}). Solutions are then selected from the sorted layers to obtain the next generation (line~\ref{ce14}-\ref{ce15}). The algorithm continues until the maximum number of generations $T$ has been reached. The final population is returned as the approximated PS and PF (line~\ref{ce17}).

\begin{algorithm}[htbp]
\small{
\caption{CE-MOEA}
\label{alg2}
\LinesNumbered
\KwIn{An attributed network ${\cal G}$, the population size $N$, the maximum number of generations: $T$; the parameters of the DE operator ($F_{DE}$ and $CR$); and the PM operator parameter: $p_m$ and $\eta_m$.}
\KwOut{an approximated PS and PF.}
Set $g \leftarrow 1$ and randomly initialize $P_g \in [0,1]^{N\times d}$;\label{ce1} \\
Evaluate $F_i \leftarrow F(P_g(i,:)), 1\leq i \leq N$; Set ${\cal F}_1 = \{F_i\}$;\label{ce2}\\
$L_s \leftarrow \text{FNS}({\cal F})$ and $C_d \leftarrow \text{CWD}({\cal F})$;\label{ce3} \\
\While{$g < T$}{ \label{ce4}
${\cal P} \leftarrow \text{BTS}(P_g, {\cal F}_g, C_d)$;\label{ce5}\\
Set ${\cal Y} = \emptyset$ and $\mathcal{F}_o = \emptyset$; \\
\For{$1\leq j\leq |\cal {P}|$}{ \label{ce6}
$\mathbf{x}_1 \leftarrow {\cal P}(j,:)$;\label{ce7} \\
Randomly select $\mathbf{x}_2$ and $\mathbf{x}_3$ from $\{{\cal P}\setminus \mathbf{x}_1\}$;\label{ce8} \\
$\mathbf{y} = \text{DE}(\mathbf{x}_1, \mathbf{x}_2, \mathbf{x}_3, F_{DE}, CR, p_m, \eta_m)$;\label{ce9} \\
${\cal Y} \leftarrow {\cal Y} \bigcup \mathbf{y} $;\\
$\mathcal{F}_o \leftarrow \mathcal{F}_o \bigcup F(\mathbf{y})$;\label{ce10} \\
}
$P_g \leftarrow P_g \cup {\cal Y}$ and ${\cal F}_g \leftarrow {\cal F}_g \cup \mathcal{F}_o $;\label{ce11} \\
$L_s \leftarrow \text{FNS}({\cal F}_g)$ and $C_d \leftarrow \text{CWD}({\cal F}_g)$;\label{ce12} \\
Sort $L_s$ based on $C_d$ in the descending order;\label{ce13} \\
Select non-dominated solutions from the sorted $L_s$ to fill $P_{g+1}$ until its size equals to $N$;\label{ce14} \\
$g\leftarrow g+1$;\label{ce15} \\
} \label{ce16}
\Return $P_T$ as the approximated PS and ${\cal F}_T$ as the PF. \label{ce17}}
\end{algorithm}

Alg.~\ref{alg3} summarizes the $\text{DE}(\cdot)$ function used to generate offsprings. In Alg.~\ref{alg3}, $\mathbf{x}_1$ is first mutated by taking the difference of $\mathbf{x}_2$ and $\mathbf{x}_3$, while the mutation takes effect only when a random number in $(0,1)$ (output by function $rand()$) is less than $CR$ (line~\ref{mutate}). The obtained $\mathbf{y}$ is repaired if any of its element is beyond the variable range (line~\ref{repair}). The PM operator~\cite{PM} is the used to mutate $\mathbf{y}$ (line~\ref{pm}). The obtained individual $\mathbf{y}$ is repaired (line~\ref{repair2}) and returned (line~\ref{yreturn}).

Advantages of the proposed approach can be summarized as follows:
1) The continuous encoding method makes fully use of the adjacency information in a network by means of the softmax layer. This can increase the robustness of the search and ensure an MOEA to have a good performance; 2) The continuous encoding method can be applied to attribute or non-attribute network,  and to undirected or directed network; 3) By transforming a discrete optimization problem into a continuous one, any promising MOEAs for continuous MOPs can be applied. As later described in the fitness landscape analysis, we find that the continuous encoding can result in a smoother landscape which are beneficial for problem solving.

\begin{algorithm}[htbp]
\small{
\caption{The Differential Evolution Operator}
\label{alg3}
\LinesNumbered
\KwIn{individuals $\mathbf{x}_1, \mathbf{x}_2$ and $\mathbf{x}_3\in \mathbb{R}^d$ and recombination parameters $F_{DE},CR,p_m$ and $\eta_m$.}
\KwOut{An offspring $\mathbf{y}$.}
\For{$1\leq i \leq d$}{
\begin{eqnarray}\nonumber
y^i=\left\{
\begin{array}{lr}
x_1^i + F_{DE} \times (x_2^i - x_3^i), \quad\text{if } rand() \leq CR,\\
x_1^i, \quad\text{otherwise} \nonumber \\
\end{array} \right.
\end{eqnarray}\label{mutate} \\
}
For $i\in \{1,...,d\}$, if $y^i < a_i$, $y^i = a_i$, otherwise, if $y^i > b_i$, $y^i = b_i$;\label{repair} \\
\For{$1\leq i \leq d$}{
\begin{eqnarray}\nonumber
y^i=\left\{
\begin{array}{lr}
y^i + \delta_i \times (b_i - a_i), \quad\text{if } rand() < p_m,\\
y^i, \quad\text{otherwise} \\
\end{array} \right.
\end{eqnarray}
and
\begin{eqnarray}\nonumber
\delta_i=\left\{
\begin{array}{lr}
[2\widehat{r}+(1-2\widehat{r})(\frac{b_i-y^i}{b_i-a_i})^{\eta_m}]^{\frac{1}{\eta_m}}-1, \quad\text{if } \widehat{r} < 0.5,\\
1\!-\![2\!-\!2\widehat{r}\!+\!(2\widehat{r}\!-\!1)(\frac{y^i-a_i}{b_i-a_i})^{\eta_m}]^{\frac{1}{\eta_m}}, \quad\text{otherwise} \\
\end{array} \right.
\end{eqnarray}where $\widehat{r}$ is an uniform random number in $[0,1]$;\label{pm} \\}
Repair $\mathbf{y}$ if necessary.\label{repair2} \\
\Return $\mathbf{y}$.\label{yreturn}
}
\end{algorithm}

\subsection{Notes}

Besides the proposed transformation method, the application of continuous approaches for solving discrete problems has also been studied based on different characterizations or reformulation. As pointed in~\cite{pardalos2006continuous}, continuous approaches may be able to reveal some new properties to the original problem. This will allow the development of new promising approaches. Researchers have tried on some typical discrete problems based on equivalent continuous formulations or relaxations, such as 0-1 programming~\cite{li2010gradient}, maximum clique problem~\cite{harant1999}, discrete DC programming~\cite{Maehara2018}, nonlinear mixed programming~\cite{Yu2013}, and others.

In these methods, to transform a discrete problem, it is required to represent its decision variables as binary. However, it is difficult to represent the community detection problem as a binary problem. To our knowledge, no continuous approaches for community detection have been studied in literature.

\subsection{Complexity Analysis}\label{c.4}

Let $r$ be the number of nodes of a network ${\cal G}$, $N$ be the population size, $m$ be the number of objectives, and $L$ be the number of edges. Alg.~\ref{alg1} requires a complexity of ${\cal O}(L)$ for the encoding process. The decoding process is the same as the locus-based decoding which requires a complexity of ${\cal O}(r)$~\cite{MOCK}. Thus, the total complexity of Alg.~\ref{alg1} is ${\cal O}(L + r)$.

For CE-MOEA, its complexity at each generation is ${\cal O}(mN^2)$ which is the same as the complexity of NSGA-II. CE-MOEA needs ${\cal O}(L + r)$ for decoding each individual and ${\cal O}(L + r)$ for evaluating each individual. CE-MOEA also requires a complexity of ${\cal O}(LN)$ as the overhead for population initialization. Overall, the total complexity of CE-MOEA is ${\cal O}(mN^2T + (L+r)NT)$.

\section{Experiment Results}\label{d}

In this section, experiments are carried out on a variety of networks with different types, with or without ground truth. CE-MOEA is implemented in Matlab 2017b on a PC. The parameter settings of CE-MOEA is as follows: the number of population is $N = 100$, the maximum number of generations is $T = 200$, the DE parameters $F_{DE} = 0.7, CR = 0.5$, the mutation probability $p_m = 0.02$ and distribution index of mutation $\eta_m = 20$. CE-MOEA was run 31 times independently in all the study.

\subsection{The Benchmark Networks}\label{d.1}

A number of networks including single- and multi-attribute networks have been used as the benchmark in our study.

Among these networks, Amazon U.S. Political Books~\cite{books} and Blogs~\cite{blog} are with single attribute. They have no truth labels for the communities. The politics books dataset include all the books studying U.S. politics which were published for presidential election and sold by Amazon.com during 2004. It contains 105 nodes and 441 edges. An edge between two books means that the two books were both purchased by customers. Each book is associated with one attribute to demonstrate political complexion: 1) conservative; 2) liberal; and 3) neutrality. The political blogs dataset was compiled by Adamic and Glance in 2005 to show the political orientation of blogs. It contains 1490 nodes and 19025 edges which connect blogs by hyperlinks. Each web-blog has an attribute showing political complexion: 1) liberal or 2) conservative.

\begin{table}[htbp]
\centering
\caption{Detailed characteristics of the benchmark networks.}
\label{table1}
\resizebox{0.95\textwidth}{!}{
\begin{tabular}{llrrrr}
\toprule
\textbf{Dataset}   & \textbf{Network Type} &\textbf{ Nodes} & \textbf{Edges} & \textbf{Attributes} & \textbf{with ground truth}\\
\midrule
Polbooks  & Books co-purchasing& 105   & 441   & 1   &    No   \\
Polblogs  & Blogs hyperlinks   & 1490  & 19025 & 1     &    No \\
Ego 0    & Friendship          & 347   & 2519  & 224        & 	No\\
Ego 107  & Friendship          & 1016  & 25711 & 576    &   No \\
Ego 686  & Friendship          & 170   & 1656  & 63        & No \\
Ego 1684 & Friendship          & 776   & 13826 & 319    &  No  \\
Ego 1912 & Friendship          & 748   & 29552 & 480     &  No \\
Ego 3437 & Friendship          & 542   & 4749  & 262      &  No \\
Ego 3980 & Friendship          & 58    & 143   & 42         &No \\
Cora      & Citation            & 2708  & 5429  & 1433      &   Yes\\
Citeseer & Citation            &3312 &4732  & 3703      & Yes  \\
Texas   &  \multirow{4}{*}{\begin{tabular}[c]{@{}l@{}}A subset networks containing \\web pages and hyperlink data\\ of the four US universities \\dataset from WebKB dataset\end{tabular}}  &  187    & 328   &1703     &  Yes \\
Cornell &     &                 195    & 304   &   1703 	& Yes\\
Washington &        &           230    & 446   & 1703	& Yes\\
Wisconsin &         &           265    & 530   &  1703 	& Yes \\
\bottomrule
\end{tabular}}
\end{table}

The rest of the networks, including the Ego facebook networks~\cite{ego} (no ground truth available), the WebKB networks~\cite{WebKB}, the Cora citation network~\cite{PCL} and the Citeseer citation network~\cite{PCL} are with multi-attributes. Ego facebook networks are a series of friendship networks. They are chosen from ten ego-networks, consisting of 4039 users. The attribute dimension of all networks ranges from 42 to 576. A subset of WebKB dataset~\cite{WebKB} consisting of four subnetworks from four U.S. universities: Cornell, Texas, Washington and Wisconsin are used. The attribute dimension of all four networks: Texas, Cornell, Washington and Wisconsin is 1703, which represent the web pages and hyperlinks between them. The Cora dataset has 2708 nodes and 5429 edges, representing the scientific publications and their citation relationships. Each publication has been divided into seven categories: 1) neural network; 2) case-based reasoning; 3) genetic algorithms; 4) probabilistic methods; 5) reinforcement learning; 6) rule learning; and 7) theory. The attribute dimension of the Cora network is 1433. The Citeseer dataset has 3312 nodes and 4732 edges. Each publication has been classified into six categories: 1) artificial intelligence; 2) database; 3) information retrieval; 4) machine learning; 5) agents; and 6) human-computer interaction. The attribute dimension of Citeseer is 3703. Table~\ref{table1} summarizes the detailed information of the benchmark networks.

\subsection{Evaluation Metrics}\label{d.2}

To compare the performances of the compared algorithms, the following metrics, including density, entropy and normalized mutual information (NMI), are used. The density and entropy metrics are applied to measure the detection performance on networks without ground truth, while NMI is for networks with true labels.

\subsubsection{Density} The density $D$ of a network is defined as
\begin{equation}
D = \sum_{k = 1}^{c}\frac{l_k}{L}\label{5}
\end{equation}where $c$ is the number of communities, $l_k$ is the number of edges within community $k$ and $L$ represents the total number of edges in the network. The larger $D$, the more distinct the community structure in the network is.

\subsubsection{Entropy} The entropy $E$ of a network is defined as
\begin{equation}\label{6}
\begin{split}
&E = \sum_{k = 1}^{c}\frac{r_k}{r}\cdot H(k)\\
&H(k) = -\sum_{a\in {\cal A}}p_{ak}\log(p_{ak})
\end{split}
\end{equation}where $p_{ak}$ is the percentage of nodes in a community $C$ with attribute value $a$. $r_k$ is the number of nodes in a cluster $k$, $c$ is the total number detected clusters, $r$ is the total number of nodes in the network. The smaller the $E$ value, the more homogeneous of nodes in the detected communities is.

\subsubsection{Normalized mutual information (NMI)} $\text{NMI}$~\cite{NMI2005} is proposed to measure the similarity between the true partitions and the detected communities. Given two partitions $P$ and $P^\ast$ of a network and $M$ be the confusion matrix whose element $M_{ij}$ is the number of nodes in community $i$ of the partition $P$ which are also in the community $j$ of partition $P^\ast$. The $\text{NMI}(P,P^\ast)$ is defined as
\begin{equation}
\text{NMI}(P,P^\ast)=\frac{-2\sum_{i=1}^{c_P}\sum_{j=1}^{c_{P^\ast}}M_{ij}\log\left( \frac{r\cdot M_{ij}}{M_{i\cdot}M_{\cdot j}}\right)}{\sum_{i=1}^{c_P}M_{i\cdot}\log\left(\frac{M_{i\cdot}}{r}\right)\!\!+\!\!\sum_{j=1}^{c_{P^\ast}}M_{\cdot j}\log\left(\frac{M_{\cdot j}}{r}\right)} \label{7}
\end{equation}where $c_P$ (resp. $c_{P^\ast}$) is the number of clusters in the partition $P$ (resp. $P^\ast$). $M_{i\cdot}$ (resp. $M_{\cdot j}$) is the summation of elements of matrix $M$ in row $i$ (column $j$).

It is obvious that if $P = P^\ast$, then $\text{NMI}(P,P^\ast) = 1$. Otherwise, if $P$ is entirely different from $P^\ast$, $\text{NMI}(P,P^\ast) = 0$. Therefore, a larger $\text{NMI}$ indicates a better quality of the detected communities, and hence a better performance of the detection algorithm.


\subsection{Results on Networks without Ground Truth}\label{e}

Both MOEA-SA and MOGA-@Net are used for comparison. Their parameter settings are set the same as in the literature. Note that in MOGA-@Net~\cite{MOGA-@Net}, three objectives regarding the network structure are employed. In our experiment, we choose to employ $Q$ which is the same as used in CE-MOEA and MOEA-SA for a fair comparison.

\subsubsection{Results on the Political Networks}\label{e.1}

Table~\ref{table21} shows the detection results of CE-MOEA, MOEA-SA and MOGA-@Net on the Political Books and Blogs, in which the maximum, minimum, average values of $D$ and $E$ are reported in columns. The standard deviations are shown in brackets. In the corresponding column, the best metric values are typeset in bold. In addition, the Wilcoxon's rank sum test at a significance level of 5\% is performed to test whether the results obtained by CE-MOEA and the compared algorithms are significantly different. In the tables, column $WR$ shows the hypothesis test results, where $\dag$, $\approx$ and $\S$ means that the result obtained by CE-MOEA is better than, similar to, and worse than the result obtained by the compared algorithms, respectively.

From Table~\ref{table21}, it is seen that CE-MOEA obtains the best results on the two political networks than MOEA-SA and MOGA-@Net, except on $D_{\max}$ for Polbooks. The Wilcoxon's rank sum tests also suggest that CE-MOEA performs significantly better than MOEA-SA and MOGA-@Net, except on the $D$ metrics for Polblogs where there is no significant difference between them. In the last column of Table~\ref{table21}, $k$ means the number of clusters obtained by the corresponding algorithm. We find that on the Polbooks network, CE-MOEA and MOEA-SA find similar number of clusters and MOGA-@Net obtains more. The communities found by CE-MOEA are with much larger number of clusters than those found by MOEA-SA and MOGA-@Net on the Polblogs.

\begin{table*}[htbp]
\centering
\caption{Statistics of the $D$ and $E$ values obtained by CE-MOEA and the compared algorithms on the political networks, where $WR$ means the Wilcoxon's Ranksum test at the 5\% significance level.}
\label{table21}
\resizebox{0.8\textwidth}{!}{
\begin{tabular}{llrrrcrrrcr}
\toprule
\textbf{Dataset}      & \textbf{Algorithms}  & $D_{\max}$ & $D_{\min}$ & $D_{\text{avg}}$& $WR$  & $E_{\max}$& $E_{\min}$ & $E_{\text{avg}}$& $WR$   & $k$    \\
\midrule
\multirow{3}{*}{Polbooks} & CE-MOEA    &0.907 & \textbf{0.889} & \textbf{0.899(0.007)}& & 0.200 & \textbf{0.091} & \textbf{0.149(0.027)}&  & 5-8  \\
& MOEA-SA  & 0.864 & 0.849 & 0.859(0.005)&$\dag$ & 0.267 & 0.206& 0.243(0.021)&$\dag$ & 4-8    \\
& MOGA-@Net  &\textbf{0.945}&0.789& 0.869(0.049)& $\dag$   &\textbf{0.198}&0.114&0.186(0.024)& $\dag$  &4-10   \\
\hline
\multirow{3}{*}{Polblogs} & CE-MOEA   &\textbf{0.916} &\textbf{0.896} & \textbf{0.906(0.006)}& & \textbf{0.042} & \textbf{0.030} & \textbf{0.036(0.004)}&  & 14-32    \\
& MOEA-SA  &0.914 & \textbf{0.896} &\textbf{0.906(0.007)}   &$\approx$ &0.153  &0.117 &0.139(0.009)  &$\dag$  & 4-16 \\
& MOGA-@Net  &0.902 &0.891 &0.898(0.003)  &$\approx$   & 0.095 &0.040 &0.061(0.013)& $\dag$  &3-8   \\
\bottomrule
\end{tabular}}
\end{table*}

\begin{figure}[htbp]
\centering
\subfigure[]{\includegraphics[width=0.45\hsize]{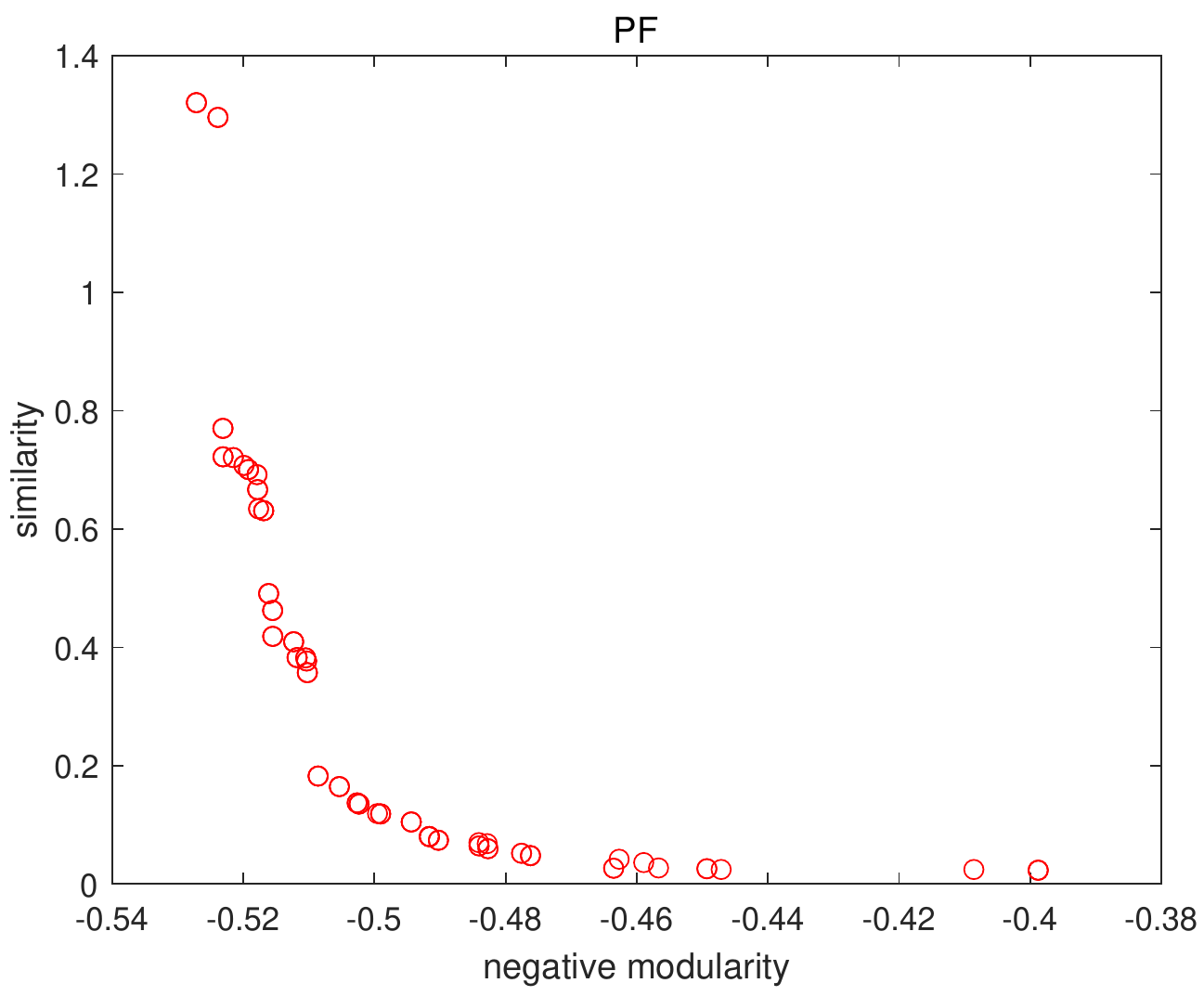}}
\subfigure[]{\includegraphics[width=0.45\hsize]{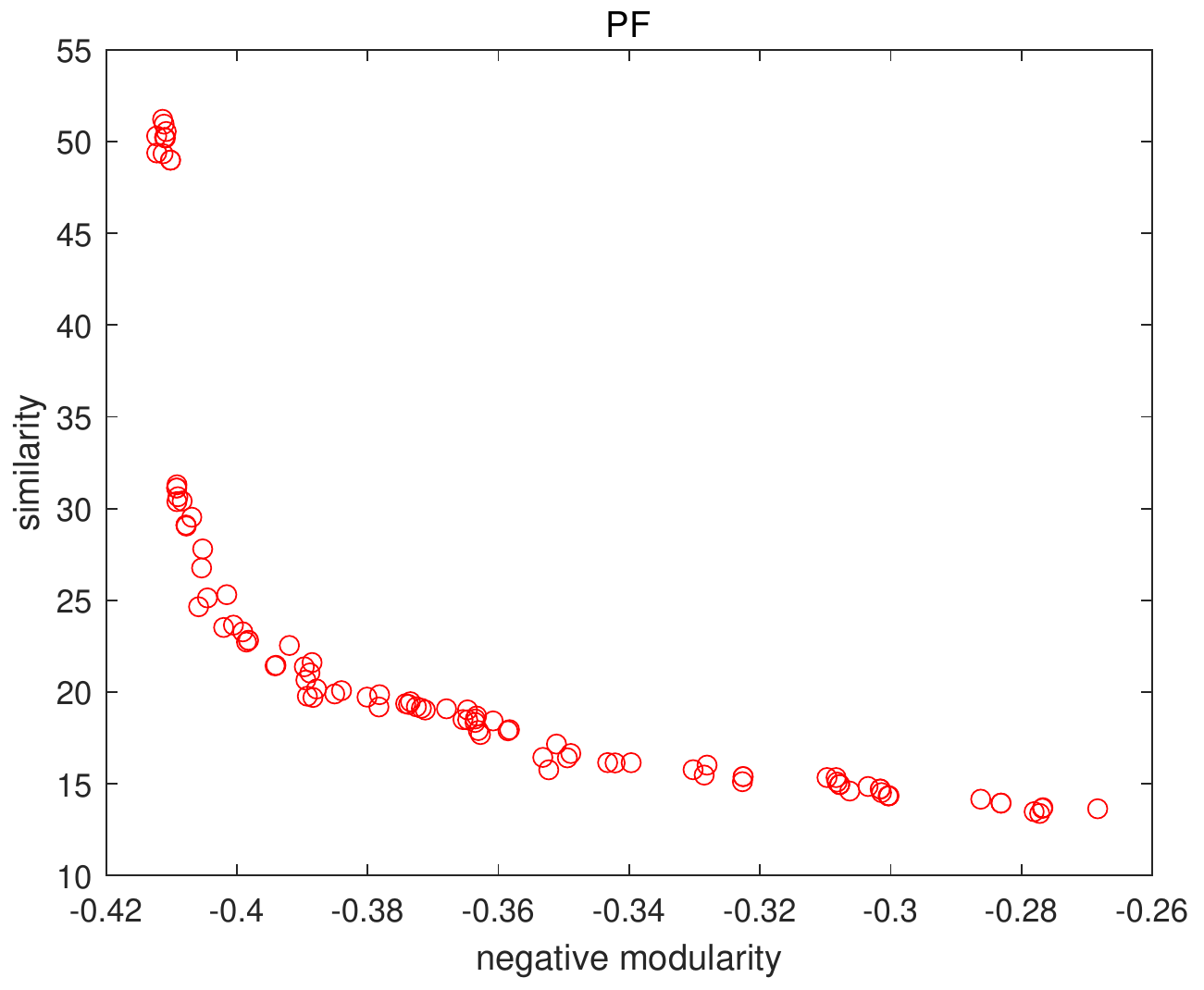}}
\caption{The PF plots of (a) Polbooks network and (b) Polblogs network obtained by CE-MOEA.}
\label{fig3}
\end{figure}

To further show the performance of CE-MOEA, the PFs obtained among the runs with the median value $D$ are shown in Fig.~\ref{fig3} for the two political networks. In the figure, the $x$-axis is the negative modularity, the $y$-axis shows the attribute similarity. From the two figures, we found that $Q$ and $f_s$ are indeed conflicting with each other. Further, it is seen that the PFs obtained by CE-MOEA are almost evenly distributed, which could reflect the good performance of CE-MOEA.

\subsubsection{Results on the Facebook Ego Networks}\label{e.2}

Experimental results of the seven ego facebook networks with multi-attribute and no ground truth are given in Table~\ref{table31}. Again, in the columns, the best metric values are marked in bold and the rank sum test results at 5\% significance level are shown.

It is seen from Table~\ref{table31} that in terms of the average $D$ and $E$, CE-MOEA always obtain better results than MOEA-SA and MOGA-@Net. The standard deviations of the obtained $D$ values are all less than 0.03, expect for Ego 3980. This clearly shows that CE-MOEA performs well and quite stable on different networks. The hypothesis test suggests that CE-MOEA performs significantly better than MOEA-SA on 5 out of 7 networks in terms of $D$. On the remaining two networks, CE-MOEA and MOEA-SA perform similarly. Table~\ref{table31} also shows that CE-MOEA performs better than MOGA-@Net on all the Ego facebook networks.

On the other hand, all the average $E$ values obtained by CE-MOEA are less than 0.171, while the standard deviations are less than 0.03. According to the hypothesis test, we found that CE-MOEA performs significantly better than MOEA-SA and MOGA-@Net on all Ego facebook networks except Ego 686 for which MOGA-@Net performs similar. Further, we found that CE-MOEA has obtained generally smaller number of communities than MOEA-SA and MOGA-@Net on all the networks, while MOGA-@Net obtains much larger  number of communities generally.

\begin{figure*}
\centering
\subfigure[]{\includegraphics[scale = 0.3]{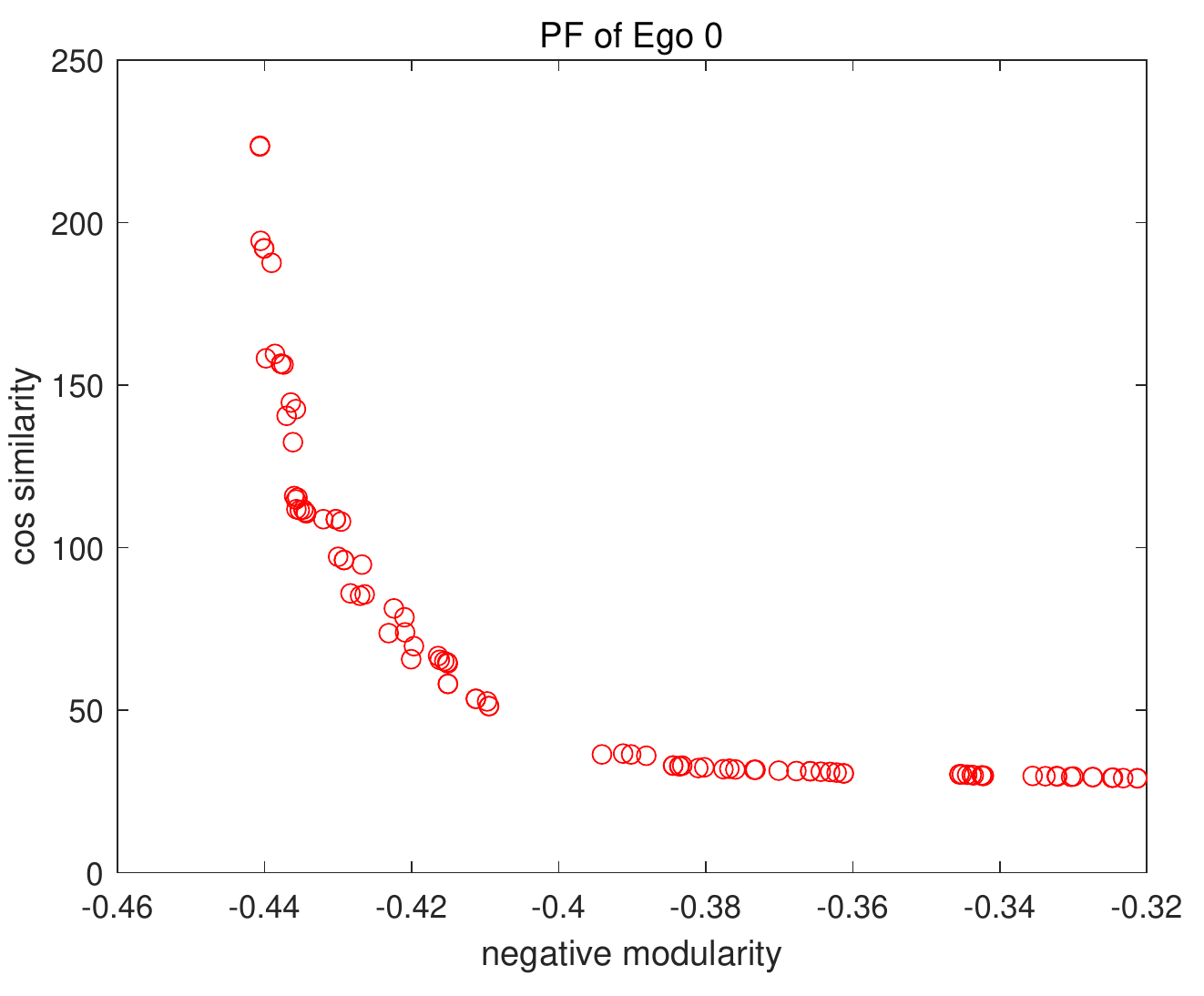}}
\subfigure[]{\includegraphics[scale = 0.3]{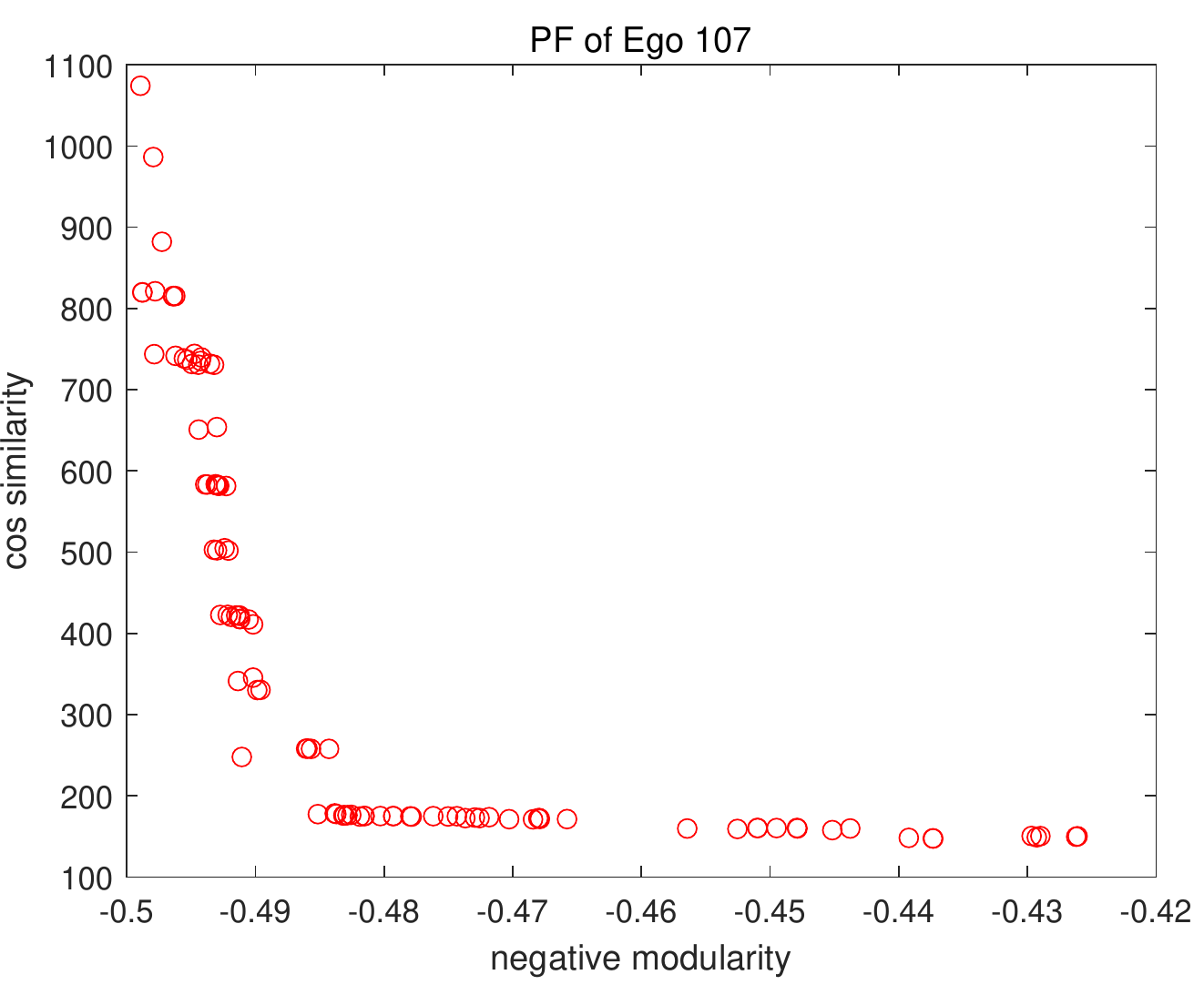}}
\subfigure[]{\includegraphics[scale = 0.3]{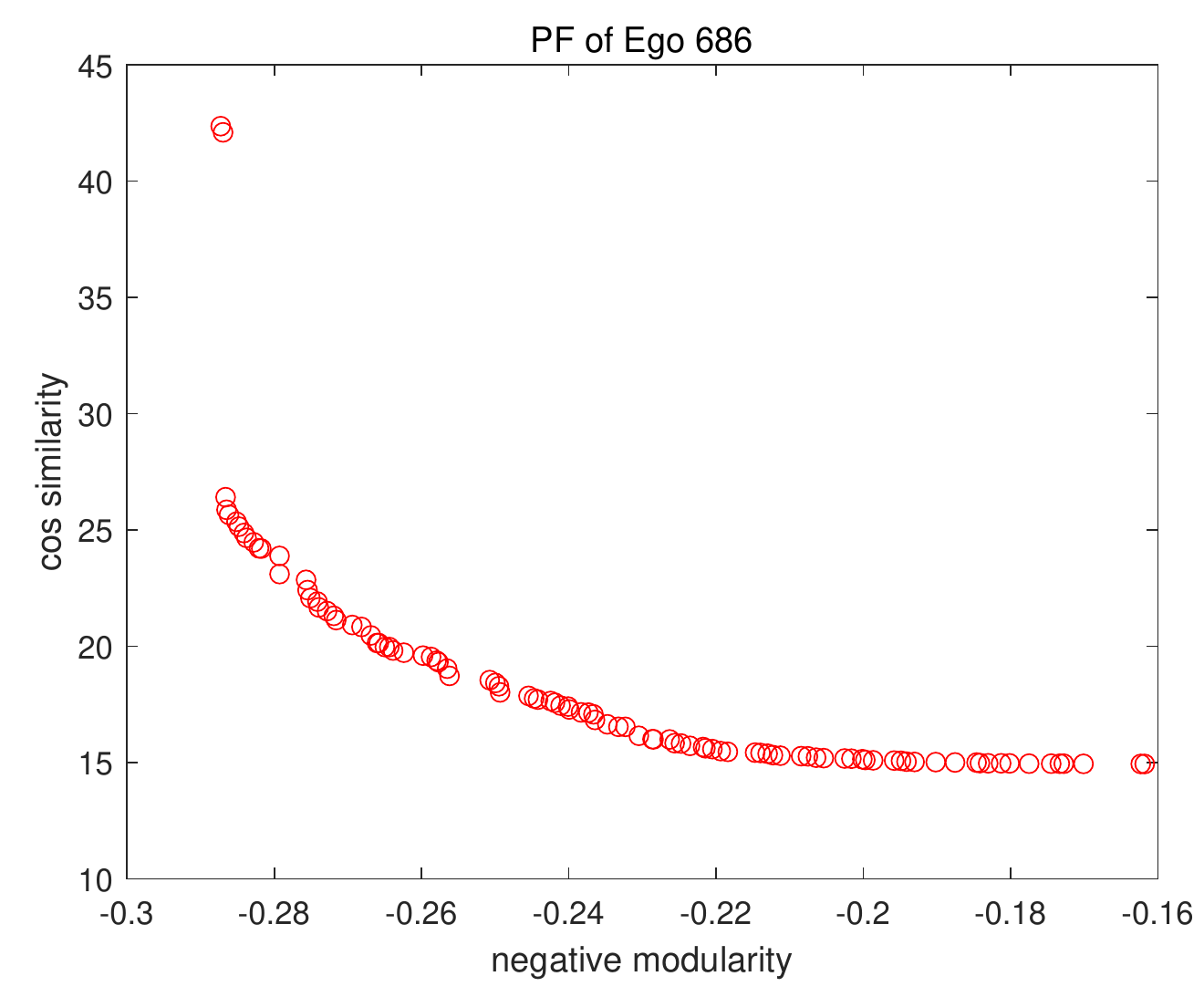}}
\subfigure[]{\includegraphics[scale = 0.3]{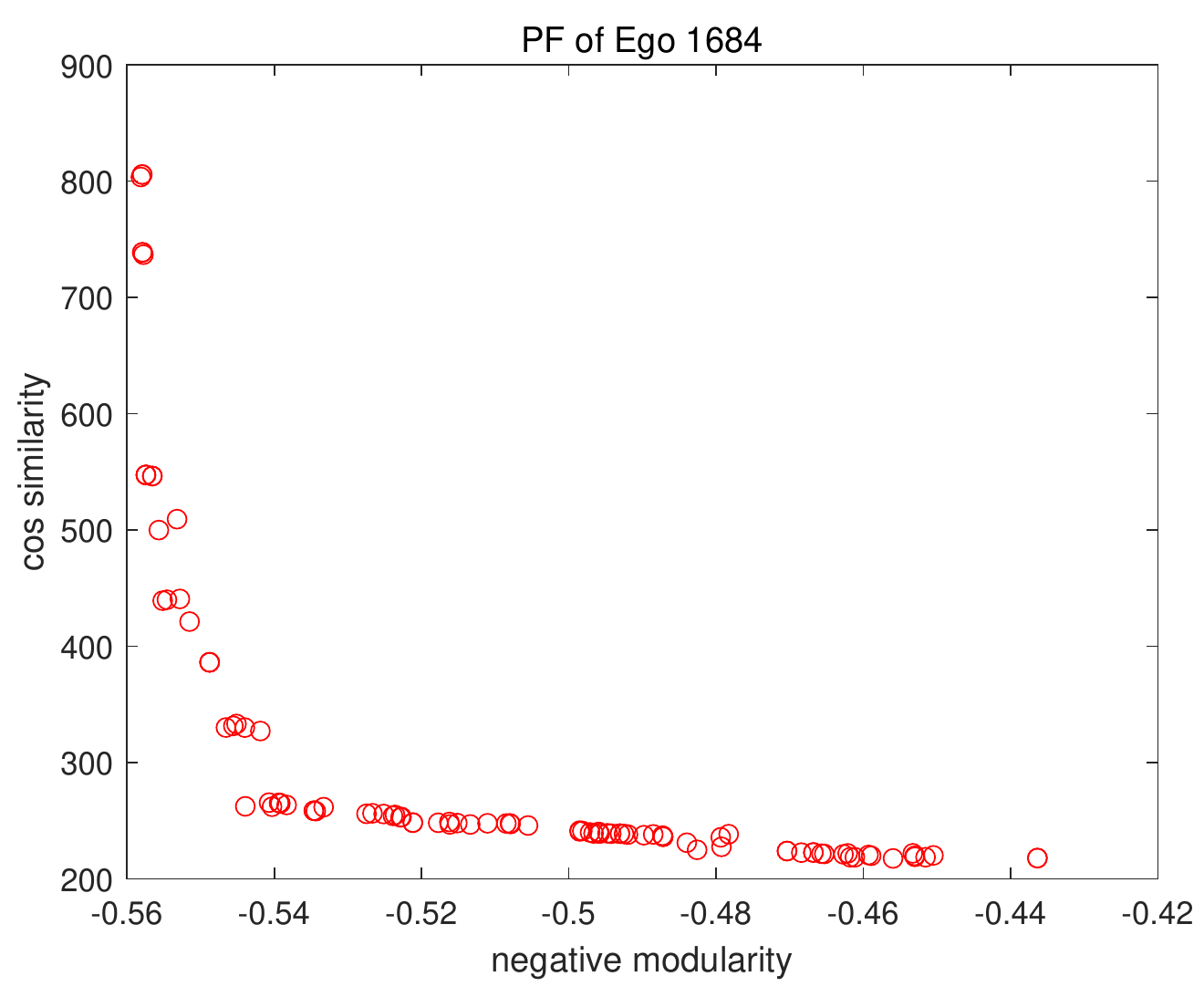}}
\subfigure[]{\includegraphics[scale = 0.3]{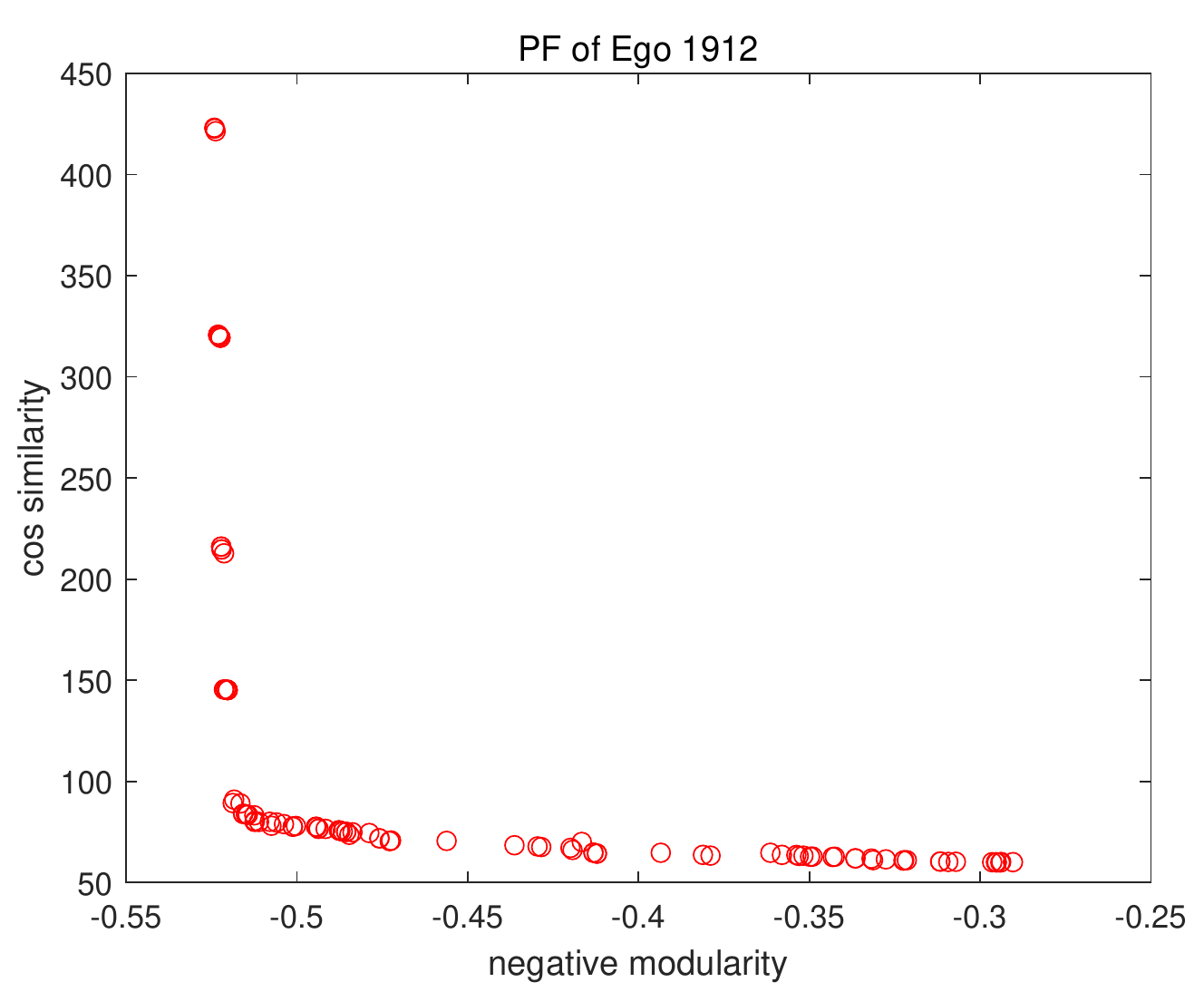}}
\subfigure[]{\includegraphics[scale = 0.3]{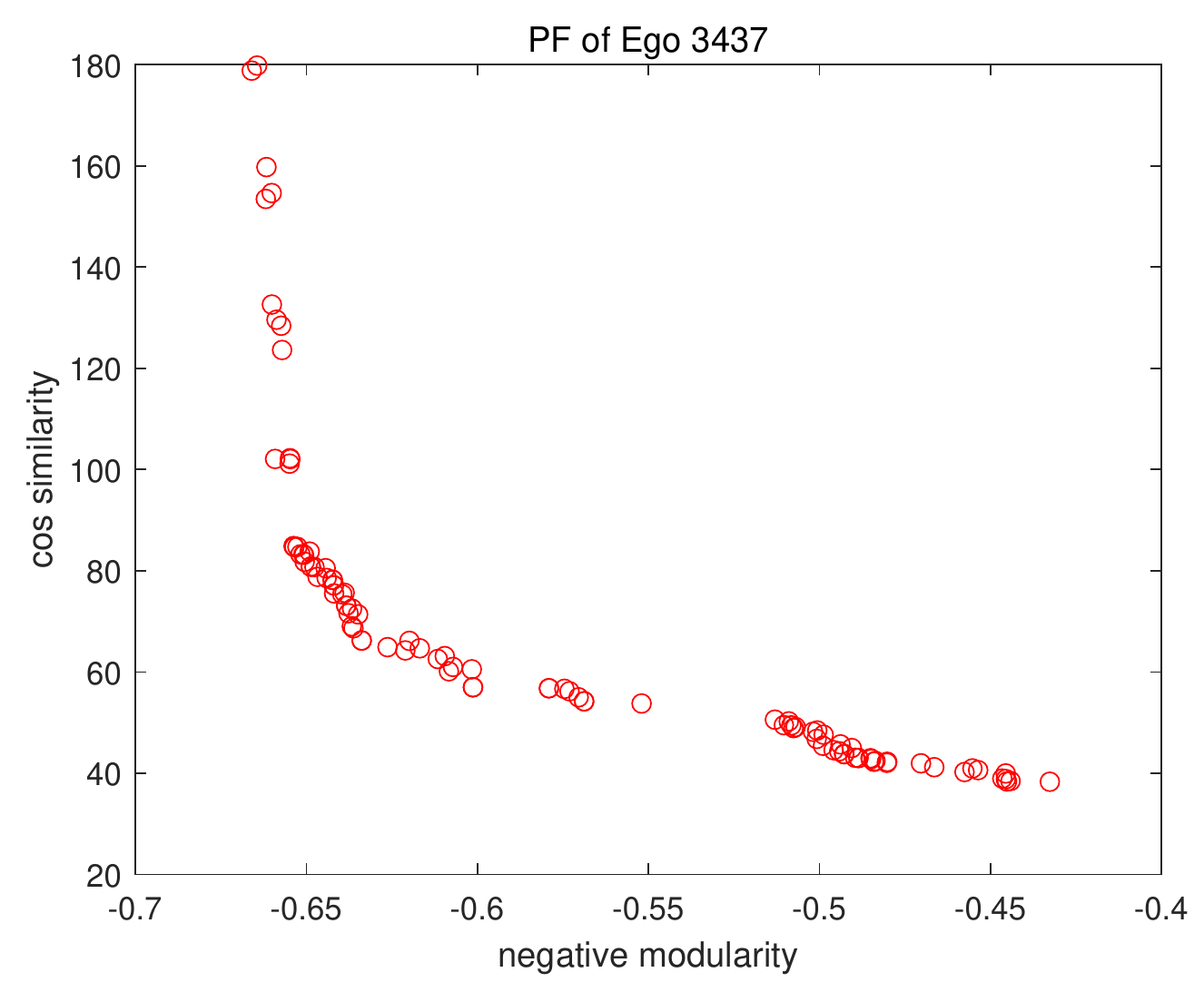}}
\subfigure[]{\includegraphics[scale = 0.3]{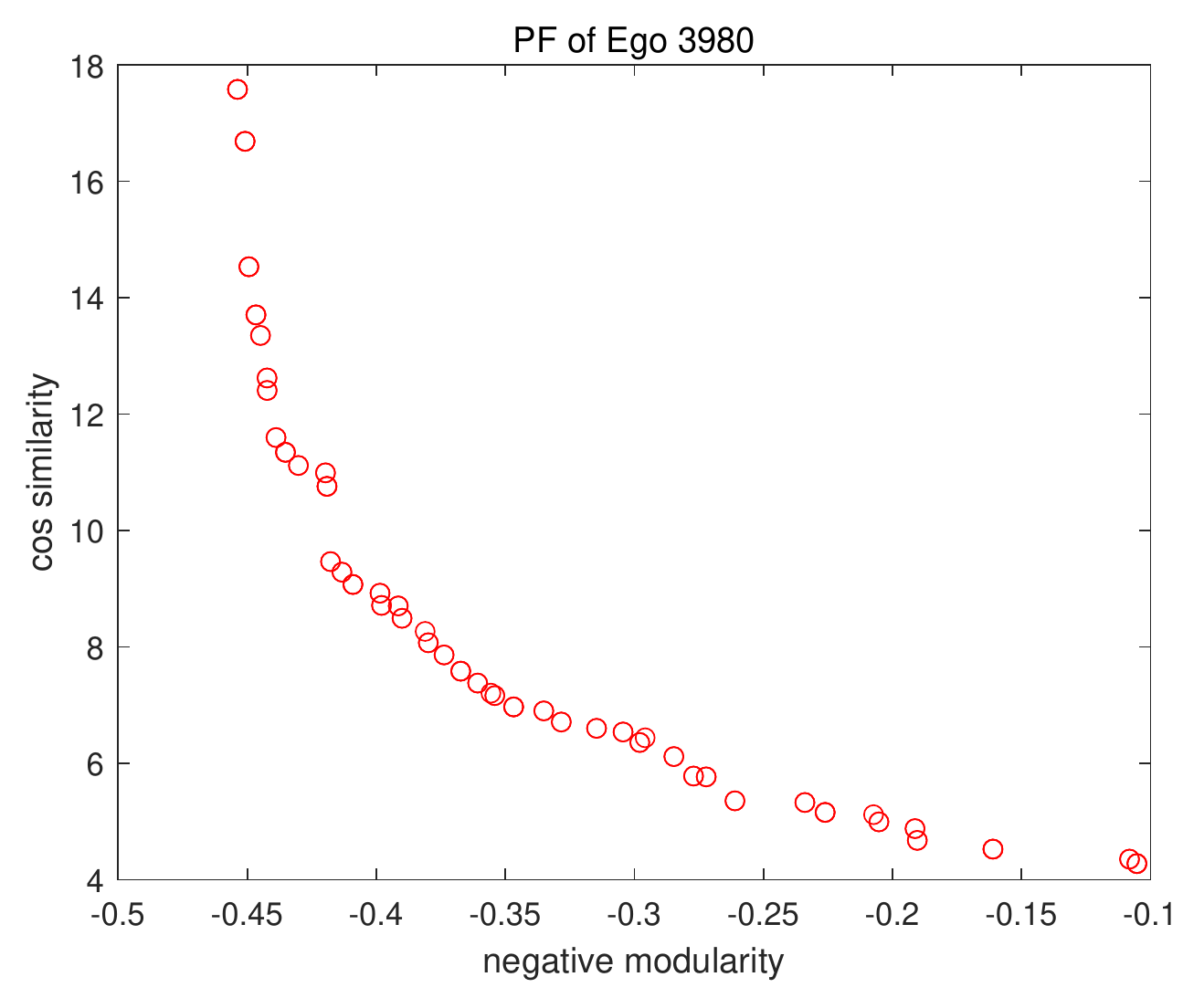}}
\caption{The PF plots of the Ego networks obtained by CE-MOEA in the run with median $D$ values.}
\label{fig4}
\end{figure*}

The PFs of the ego networks obtained by CE-MOEA in the run with the median $D$ value are shown in Fig.~\ref{fig4}. Similar to Fig.~\ref{fig3}, we find that the two objectives are conflicting with each other. Further, it is found that the PFs are mostly evenly distributed, which reflects a good performance of CE-MOEA.

In summary, we may conclude that CE-MOEA is able to achieve a good balance between network structure and attribute similarity when solving the detection problem of multi-attribute networks, and performs better than MOEA-SA and MOGA-@Net.

\begin{table*}[htbp]
\centering
\caption{Statistics of the obtained $D$ and $E$ metric values by CE-MOEA and the compared algorithms on ego facebook networks, where $WR$ means the Wilcoxon's Ranksum test at the 5\% significance level.}
\label{table31}
\resizebox{0.8\textwidth}{!}{
\begin{tabular}{llrrrcrrrcr}
\toprule
\textbf{Dataset}    & \textbf{Algorithms}  & $D_{\max}$ & $D_{\min}$ & $D_{\text{avg}}$& $WR$ & $E_{\max}$& $E_{\min}$ & $E_{\text{avg}} $  &$WR$ & $k$   \\
\midrule
\multirow{3}{*}{Ego 0} & CE-MOEA  & \textbf{0.964} &\textbf{0.859} &\textbf{0.934(0.030)}& &\textbf{0.071} &\textbf{0.040}  & \textbf{0.051(0.010)} & & 3-13  \\
         & MOEA-SA & 0.815 & 0.632 &0.707(0.049)&$\dag$ & 0.146 & 0.142 &0.144(0.001) &$\dag$ & 6-17     \\
         & MOGA-@Net &0.944 &0.642 &0.742(0.132)&$\dag$  &0.125 &0.097  &0.111(0.008) &$\dag$ & 32-34    \\
\hline
\multirow{3}{*}{Ego 107} & CE-MOEA &\textbf{0.946}&\textbf{0.930}& \textbf{0.940(0.005)}&  &\textbf{0.036} &\textbf{0.024}&\textbf{0.030(0.003)}& & 5-17 \\
& MOEA-SA   & 0.938 & 0.804 & 0.917(0.037)&$\dag$ & 0.078 & 0.076&0.077(0.001) &$\dag$ & 4-29 \\
& MOGA-@Net &0.916  &0.909  &0.911(0.002)& $\dag$&0.065 &0.057&0.061(0.002)&$\dag$ &35-38     \\
\hline
\multirow{3}{*}{Ego 686}  & CE-MOEA &\textbf{0.758}& \textbf{0.687}&\textbf{0.723(0.021)}& &\textbf{0.081} &0.067 &\textbf{0.069(0.003)}& & 3-5    \\
& MOEA-SA                  & 0.648 & 0.578 & 0.621(0.021)&$\dag$ &0.295 &0.271 &0.282(0.007) &$\dag$&4-11   \\
& MOGA-@Net &0.643  &0.435  &0.540(0.061) &$\dag$ &0.084  &\textbf{0.064}  &0.075(0.006)&$\approx$ &5-9    \\
\hline
\multirow{3}{*}{Ego 1684} & CE-MOEA & 0.900& \textbf{0.891}&\textbf{0.897(0.002)}& &\textbf{0.031}&\textbf{0.023}&\textbf{0.026(0.002)}& &5-10   \\
& MOEA-SA  & \textbf{0.926} & 0.853 & 0.888(0.019)&$\approx$ & 0.091 & 0.088 & 0.090(0.001)& $\dag$  &6-24  \\
& MOGA-@Net &0.864&0.786&0.813(0.028)&$\dag$ &0.064 & 0.056 &0.061(0.003) &$\dag$ & 26-28    \\
\hline
\multirow{3}{*}{Ego 1912} & CE-MOEA &\textbf{0.976} & \textbf{0.960} &\textbf{0.966(0.005)}& &\textbf{0.031} & \textbf{0.021} & \textbf{0.026(0.003)}& & 4-9     \\
& MOEA-SA  & 0.918 & 0.785 & 0.849(0.040)&$\dag$ & 0.090 & 0.086 & 0.087(0.001)&$\dag$&3-15  \\
& MOGA-@Net & 0.952 &0.912  &0.921(0.016)& $\dag$ &0.044 &0.039 & 0.043(0.002) & $\dag$ &18-20\\
\hline
\multirow{3}{*}{Ego 3437}& CE-MOEA  & \textbf{0.916} & \textbf{0.866} &\textbf{0.875(0.009)}& & \textbf{0.065} &\textbf{0.045} & \textbf{0.055(0.004)}& & 7-13    \\
& MOEA-SA   & 0.898 & 0.806 & 0.861(0.029)&$\approx$ & 0.106 & 0.101 & 0.103(0.001)&$\dag$ &11-23 \\
& MOGA-@Net & 0.900 & 0.836 & 0.852(0.022)&$\dag$ &0.106 &0.094  &0.103(0.003) & $\dag$ &22-25     \\
\hline
\multirow{3}{*}{Ego 3980} & CE-MOEA   & \textbf{0.923} & \textbf{0.769} & \textbf{0.855(0.048)}&  & \textbf{0.171} & \textbf{0.070}  & \textbf{0.118(0.021)}  & & 3-6    \\
& MOEA-SA  & 0.669 & 0.597 & 0.626(0.021)& $\dag$  & 0.310 & 0.267  & 0.284(0.012) &$\dag$ &6-10  \\
& MOGA-@Net &0.788 &0.709 & 0.752(0.029)& $\dag$ &0.207 &0.185  &0.194(0.009) & $\dag$ &15-17     \\
\bottomrule
\end{tabular}}
\end{table*}

\subsection{Results on Networks with Ground Truth}\label{f}

In this section, six multi-attribute networks (Cornell, Texas, Washington, Wisconsin, Cora, and Citeseer) with true labels are used as the benchmark. Ten state-of-the-art community detection algorithms including distance-based methods (SA-Cluster~\cite{SA-Cluster} and Inc-Cluster~\cite{Inc-Cluster}), model-based methods (vGraph~\cite{vGraph}, PCL~\cite{PCL}, BAGC~\cite{BAGC}, SCI~\cite{SCI},  TLSC~\cite{TLSC} and CDE~\cite{CDE}) and MOEA-based methods (MOEA-SA~\cite{MOEA-SA} and MOGA-@Net~\cite{MOGA-@Net}) are compared with CE-MOEA in terms of NMI. Note vGraph is not applicable for attribute network since it only considers network structure.

The statistics of the NMI metrics, including the mean and standard deviation, obtained by the compared algorithms are shown in Table~\ref{table4}. For each network, the best mean NMI values are typeset in bold. Further, the $z$-test was carried out to find out the differences between CE-MOEA and the compared algorithms. In Table~\ref{table4}, symbols `$+$' (resp. `$-$' and `$=$') denotes that the performance of the compared algorithm is significantly better than (resp. worse than and similar to) CE-MOEA at the 5\% significant level.

From Table~\ref{table4}, the $z$-test suggests that CE-MOEA performs significantly better than vGraph, SA-Cluster, Inc-Cluster, BAGC, TLSC, MOEA-SA and MOGA-@Net on all the networks. CE-MOEA performs worse than PCL on Cora, worse than SCI on Texas, worse than CDE on Cornell, Texas, Washington and Wisconsin. However, we find that CE-MOEA is ranked the second on Cornell, Washington, Wisconsin and Cora, ranked the third on Texas. CE-MOEA performs similarly to SCI on Wisconsin, to CDE on Cornell and Texas.

Note that except MOEA-SA, MOGA-@Net and CE-MOEA, all the other compared algorithms require to know the true number of communities as {\em a priori}. Hence, CE-MOEA is more favorable for practical application.

\begin{table*}[!t]
\centering
\caption{The mean and  standard deviation of the obtained NMI values by the compared method and CE-MOEA on multi-attribute networks with ground truth.}
\label{table4}
\resizebox{0.95\textwidth}{!}{
\begin{tabular}{l|cc|cccccc|ccc}
\toprule
\multirow{3}{*}{\textbf{Dataset}} & \multicolumn{10}{c}{\textbf{Methods}}        \\ \cline{2-12}
           & \multicolumn{2}{c}{\textbf{Distance-based}} & \multicolumn{6}{c}{\textbf{Model-based}}& \multicolumn{3}{c}{\textbf{MOEA-based}} \\ \cline{2-12}
           & SA-Cluster &Inc-Cluster &vGraph &PCL &BAGC &SCI & TLSC &CDE& MOEA-SA& MOGA-@Net & CE-MOEA \\  \midrule
Cornell&0.064$^-$ &0.038$^-$&0.039(0.012)$^-$ &0.073(0.010)$^-$ &0.040(0.006)$^-$&0.166(0.008)$^-$&0.092(0.032)$^-$&\textbf{0.203(0.043)}$^=$& 0.133(0.015)$^-$&0.172(0.016)$^-$ &0.199(0.007)  \\
Texas    & 0.082$^-$ &0.106$^-$&0.024(0.006)$^-$&0.061(0.011)$^-$ &0.052(0.007)$^-$&\textbf{0.200(0.019)}$^+$&0.122(0.063)$^-$&0.181(0.082)$^=$& 0.114(0.013)$^-$&0.091(0.020)$^-$&0.163(0.012)\\
Washington & 0.077$^-$ &0.063$^-$&0.020(0.007)$^-$&0.092(0.015)$^-$ &0.053(0.006)$^-$&0.146(0.006)$^-$&0.166(0.056)$^-$&\textbf{0.276(0.072)}$^+$& 0.140(0.013)$^-$&0.167(0.010)$^-$  &0.244(0.006) \\
Wisconsin  & 0.101$^-$&0.089$^-$&0.043(0.008)$^-$&0.060(0.001)$^-$ &0.034(0.015)$^-$&0.184(0.001)$^=$&0.081(0.022)$^-$&\textbf{0.264(0.082)}$^+$& 0.158(0.015)$^-$&0.173(0.010)$^-$ &0.191(0.006) \\
Cora       & 0.117$^-$ &0.112$^-$&0.080(0.012)$^-$&\textbf{0.416(0.003)}$^+$ &0.008(0.005)$^-$&0.205(0.008)$^-$ &0.248(0.004)$^-$&0.325(0.084)$^-$& 0.119(0.002)$^-$&0.379(0.005)$^-$&0.399(0.003)  \\
Citeseer   & 0.047$^-$ &0.043$^-$&0.052(0.012)$^-$&0.170(0.003)$^-$ &0.017(0.001)$^-$&0.077(0.037)$^-$ &0.150(0.019)$^-$&0.281(0.055)$^-$& 0.143(0.003)$^-$&0.331(0.002)$^-$&\textbf{0.346(0.001)} \\
\bottomrule
\end{tabular}}
\end{table*}

\subsection{More Experiments}\label{e.4}

To further investigate the performance of CE-MOEA, we generated three large synthesis networks with single attribute, named as LFR3500, LFR3700 and LFR3900, by using the LFR benchmark generator~\cite{LFR}. The number of nodes of these networks are 3500, 3700 and 3900 and each node has an attribute value within [1,83], [1,86] and [1,89] respectively. Further, a subnetwork (id. 629863) of the Twitter network~\cite{snapnets} is used for comparison. It has 171 nodes indicating Tweets and 796 edges and its attribute dimension is 578. Since the ground truths of these networks are known, $\text{NMI}$ is applied to evaluate the compared methods.

The following four algorithms, namely MOGA-@Net~\cite{MOGA-@Net}, MOEA-SA~\cite{MOEA-SA}, CDE~\cite{CDE} and TLSC~\cite{TLSC} are compared with CE-MOEA. The mean and standard deviations (in brackets) of the obtained $\text{NMI}$ values in 31 runs are summarized in Table~\ref{tablelarge}. From the table, we find that CE-MOEA obtains higher NMI values than those obtained by the compared algorithms in general.

\begin{table}[htbp]
\centering
\caption{The statistics of the NMI results obtained by CE-MOEA and the compared methods on Twitter and three large LFR networks.}
\label{tablelarge}
\resizebox{0.8\textwidth}{!}{
\begin{tabular}{lrrrrr}
\toprule
\textbf{Dataset}  & CE-MOEA      & MOGA-@Net & MOEA-SA      & CDE          & TLSC         \\ \midrule
Twitter & \textbf{0.473(0.020)}& 0.441(0.028)&0.470(0.034)& 0.352(0.090)   &0.325(0.051) \\
LFR3500 & \textbf{0.807(0.002)}& 0.802(0.006)& 0.798(0.006) & 0.630(0.059) &0.541(0.019)      \\
LFR3700 & \textbf{0.803(0.002)}& 0.797(0.008)& 0.796(0.009) & 0.642(0.057) &0.619(0.023)      \\
LFR3900 & \textbf{0.809(0.001)}& 0.804(0.008)& 0.798(0.005) & 0.636(0.054) &0.579(0.007)     \\
\bottomrule
\end{tabular}}
\end{table}

\subsection{Effectiveness of the proposed objectives}\label{e.5}

To verify the difference between the proposed objective function $f_s$ (or $f_m$) (cf. Eq.~\ref{3} and~\ref{4}) and $S_A$ as proposed in~\cite{MOEA-SA}, we replaced $S_A$ with $f_s$ (or $f_m$) in CE-MOEA, named as CE-MOEA-v2.

\begin{table}[htbp]
\centering
\caption{The best $E$ of the six versions of CE-MOEA on three networks.}
\label{tableobj}
\resizebox{0.7\textwidth}{!}{
\begin{tabular}{lrrrrrr}
\toprule
\textbf{Dataset} & CE-MOEA & -v2    & -v3    & -v4    & -v5    & -v6    \\ \midrule
Polbooks & \textbf{0.091}   & 0.114 & 0.116 & 0.116 & 0.116 & 0.117 \\
Ego 686  & \textbf{0.067}   & 0.070 & 0.068 & 0.068 & 0.068 & 0.071 \\
Ego 3980 & \textbf{0.070}   & 0.113 & 0.113 & \textbf{0.070} & 0.113 & 0.182 \\
\bottomrule
\end{tabular}}
\end{table}

Further, we investigated the influences of the denominator used in $f_s$ (or $f_m$). Four variants of the denominator, namely $\sum_{k = 1}^{c} r_k$, $\sum_{k = 1}^{c} r_k^2$, $\sum_{k = 1}^{c} (r_k-1)^2$ and no denominator, are incorporated within the objective, where $c$ is the number of obtained communities and $r_k$ stands for the number of nodes in the $k$-th community. The corresponding CE-MOEAs are named as CE-MOEA-v3, CE-MOEA-v4, CE-MOEA-v5 and CE-MOEA-v6, respectively. Three datasets (Polbooks, Ego 686 and Ego 3980) are used as the benchmark.

The best $E$ values are used to measure the objectives' efficacy. Table~\ref{tableobj} summarizes the results. From the table, we find that CE-MOEA performs the best especially for Polbooks, which indicates that using $f_s$ or $f_m$ are more appropriate than the other objectives.

\subsection{Comparison between MOEA-SA and CE-MOEA in terms of Running Time}\label{e.6}

We compare the average running time of one generation of CE-MOEA and MOEA-SA on three networks (including Polbooks, Ego 686 and Ego 3980). The average running times (in seconds) are summarized in Table~\ref{timetable}. From the table, we find that MOEA-SA is faster than CE-MOEA. However, note that MOEA-SA is implemented in C++ and CE-MOEA in Matlab. The difference in terms of the running time between the two algorithms is not as big as shown in the table. The running time of CE-MOEA should be acceptable for real applications.

\begin{table}[htbp]
\centering
\caption{The running time comparison between CE-MOEA and MOEA-SA for one generation (in seconds).}
\label{timetable}
\resizebox{0.4\textwidth}{!}{
\begin{tabular}{lrr}
\toprule
Dataset  & CE-MOEA      & MOEA-SA \\ \midrule
Polbooks &1.005s &0.033s \\
Ego 686  &6.475s &1.348s  \\
Ego 3980 &1.425s  &0.037s   \\
\bottomrule
\end{tabular}}
\end{table}

\section{Fitness Landscape Analysis}\label{g}

From the above experimental study, we may conclude that the proposed algorithm performs better than existing algorithms for both single- and multi-attribute networks with known or unknown ground truth.

Notice that MOEA-SA is also built upon NSGA-II. This makes us to think that maybe the proposed graph neural network encoding is the reason for the better performance of CE-MOEA. Since through graph neural network encoding, the original discrete optimization problem is transformed to a continuous one. We thus further conjecture that due to the continuous encoding, the fitness landscape of the original problem becomes smoother.

In this section, we resort to the fitness landscape analysis to confirm our conjecture. Six networks, including Polbooks, Ego 0, Ego 107, Ego 686, Ego 3437 and Ego 3980, are used as examples to conduct the analysis based on the modularity $Q$ and the attribute similarity $f_s$ or $f_m$. In our experiments, the ruggedness of the community detection problem landscape is measured by three metrics, including local optimum density (LOD), escaping rate (ER) and fitness distance correlation (FDC)~\cite{Merz2004}. All these metrics are obtained by applying the Iterated Local Search (ILS)~\cite{ILS} heuristic. The metrics are defined as follows:
\begin{itemize}
  \item LOD:  It is the number of local optima encountered by an ILS per 100 moves. Here, one move indicates that the local search moves from the current solution to a new solution within its neighborhoods.
  \item ER: This refers to the success rate of the ILS to reach a new local optimum by perturbing the current local optimum.
  \item FDC: To compute FDC, we have randomly selected 1000 local optima ($\mathbf{x}_{LO}$) from the set obtained by the ILS and their function values are $f(\mathbf{x}_{LO})$. Then, the distances of 1000 local optima to the nearest global optimum are calculated as $d_{\text{opt}}$. Overall, the FDC is defined as
\begin{equation} \text{FDC}(f(\mathbf{x}_{LO}),d_{\text{opt}}) = \frac{\texttt{cov}(f(\mathbf{x}_{LO},d_{\text{opt}}))}{\sigma(\mathbf{x}_{LO})\sigma(d_{\text{opt}})}
\end{equation}where $\texttt{cov}(\cdot)$ means the covariance and $\sigma(\cdot)$ means the standard deviation.
\end{itemize}
It is generally acknowledged that a lower LOD (and ER) or a higher FDC means that a heuristic can find the global optimum easier, which means a smoother landscapes~\cite{FDC}.
\begin{table}[htbp]
\centering
\caption{The fitness landscape metrics obtained for the original problem and transformed problem on the selected networks in terms of the modularity $Q$.}
\label{table5}
\resizebox{0.70\textwidth}{!}{
\begin{tabular}{lrrrrrr}
\toprule
\multirow{2}{*}{\textbf{Dataset}} & \multicolumn{6}{c}{\textbf{Metrics}}                                                             \\ \cline{2-7}
                          &$\texttt{LOD}_{o}$&$\texttt{LOD}_{t}$&$\texttt{ER}_{o}$ &$\texttt{ER}_{t}$             &$\texttt{FDC}_{o}$&$\texttt{FDC}_{t}$  \\ \midrule
Polbooks                  &4.562 &\textbf{4.282}&0.519 &\textbf{0.001} &0.048 &\textbf{0.133} \\
Ego 0                     &3.821 &\textbf{3.756}&0.561 &\textbf{0.034} &0.168 &\textbf{0.189}   \\
Ego 107                   &4.211 &\textbf{2.033}&0.590 &\textbf{0.027} &0.291 &\textbf{0.326}   \\
Ego 686                   &4.220 &\textbf{1.982}&0.535 &\textbf{0.003} &0.172 &\textbf{0.195}    \\
Ego 3437                  &4.290 &\textbf{2.861}&0.019 &\textbf{0.009} &0.102&\textbf{0.128}     \\
Ego 3980                  &3.421 &\textbf{3.235}&0.510 &\textbf{0.004} &0.023 &\textbf{0.217}     \\
\bottomrule
\end{tabular}}
\end{table}
\begin{table}[htbp]
\centering
\caption{The fitness landscape metrics obtained for the original problems and transformed problems on the selected networks in terms of $f_s$ or $f_m$.}
\label{table6}
\resizebox{0.70\textwidth}{!}{
\begin{tabular}{lrrrrrr}
\toprule
\multirow{2}{*}{\textbf{Dataset}} & \multicolumn{6}{c}{\textbf{Metrics}}                                                             \\ \cline{2-7}
                          &$\texttt{LOD}_{o}$&$\texttt{LOD}_{t}$&$\texttt{ER}_{o}$ &$\texttt{ER}_{t}$             &$\texttt{FDC}_{o}$&$\texttt{FDC}_{t}$  \\ \midrule
Polbooks                 &5.359 &\textbf{4.217}&0.048 &\textbf{0.002} &0.067 &\textbf{0.209} \\
Ego 0                     &3.951 &\textbf{3.789}&0.050 &\textbf{0.001} &0.147 &\textbf{0.219}   \\
Ego 107                 &4.208 &\textbf{4.142}&0.517 &\textbf{0.513} &0.136 &\textbf{0.141}   \\
Ego 686                 &4.214&\textbf{3.929}&0.003 &\textbf{0.001} &0.141 &\textbf{0.151}    \\
Ego 3437               &4.197&\textbf{3.979}&0.005 &\textbf{0.004} &0.138 &\textbf{0.158}     \\
Ego 3980               &4.206&\textbf{3.255}&0.004 &\textbf{0.002} &0.213 &\textbf{0.317}     \\
\bottomrule
\end{tabular}}
\end{table}

The ILS performs a local search process and a perturbation process iteratively until the stopping condition is met. In the local search process, it tries to find a better solution in current solution's neighborhood. If such a solution is found, it is used to replace the current solution. The process continues until there is no better solution in the neighborhood. A perturbation is performed once the local search process is stuck.

In our study, the fitness landscape analysis is based on the locus-based encoding method for the original problem. The neighborhood is defined as follows. Given two genotypes $\mathbf{x} = (x^1, x^2, \ldots, x^r)$ and $\mathbf{y} = (y^1, y^2, \ldots, y^r)$, the distance between them is defined by
\begin{equation}
 \text{dist}(\mathbf{x},\mathbf{y}) = \sum_{i=1}^{r}|\textrm{sgn}(x^i - y^i)|
\end{equation}
where
\begin{equation}
\textrm{sgn}(z) = \left\{
\begin{array}{lr}
1, \hspace{5mm}z \neq 0,\\
0, \hspace{5mm}z = 0.
\end{array}\right.
\end{equation}
The neighborhood of a genotype $\mathbf{x}$ is thus defined as
\begin{equation}
{\cal N}_{O}(\mathbf{x}) = \{\mathbf{y}|\text{dist}(\mathbf{x},\mathbf{y}) = 1\}
\end{equation}The perturbation process is implemented by replacing ten random edge of the current solution.

For the transformed problem, the neighborhood of a solution $\mathbf{x}$ is defined as
\begin{equation}
{\cal N}_{T}(\mathbf{x};\epsilon) = \{\mathbf{x}'|\|\mathbf{x}-\mathbf{x}'\|_2\leq \epsilon\}
\end{equation}where $\epsilon$ is a threshold. For a certain problem, $\epsilon$ is obtained as follows: firstly, we sample 100,000 solutions randomly. The maximum distance $d_{\max}$ and the minimum distance $d_{\min}$ among the solution pairs are used to compute $\epsilon = (d_{\max} + d_{\min})/2$. To apply the ILS, in the perturbation process, we randomly sample a solution such that its distance to the current solution is greater than $\epsilon$.

To carry out fitness landscape analysis, we first obtain a set of 10,000 local optima by applying the ILS method. Based on the obtained local optima, the fitness landscape metrics are computed, which are shown in Tables~\ref{table5} and~\ref{table6} for modularity and attribute similarity, respectively. In the tables, $\texttt{LOD}_{o}$ (resp. $\texttt{ER}_{o}$ and $\texttt{FDC}_{o}$) means LOD (resp. ER and FDC) metric for the original problem and $\texttt{LOD}_{t}$ (resp.  $\texttt{ER}_{t}$ and $\texttt{FDC}_{t}$) is for the transformed problem. The better results of the corresponding metrics are typeset in bold.
\begin{figure}[H]
\centering
\subfigure[]{\includegraphics[width=0.24\linewidth,height=2cm]{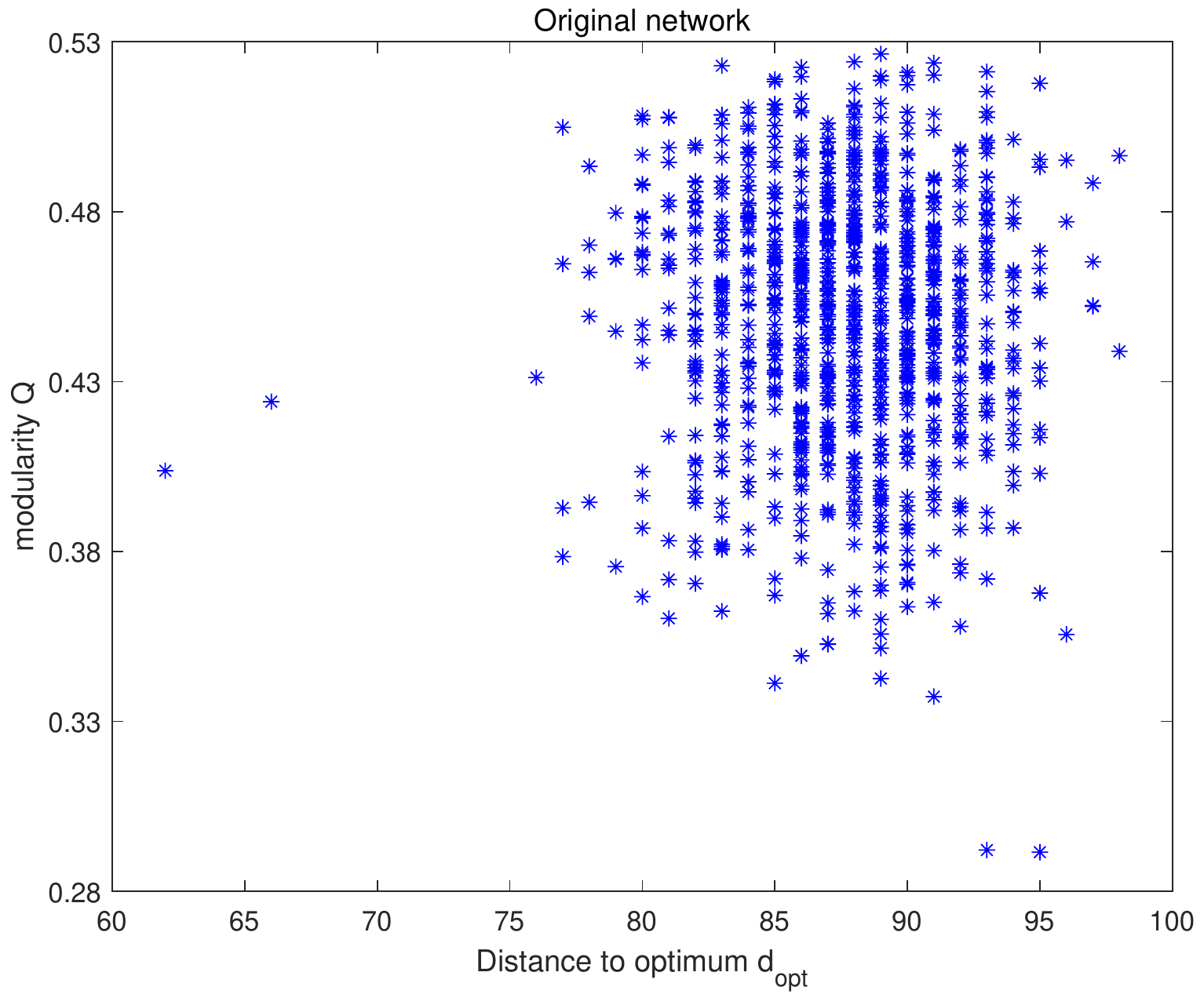}}
\subfigure[]{\includegraphics[width=0.24\linewidth,height=2cm]{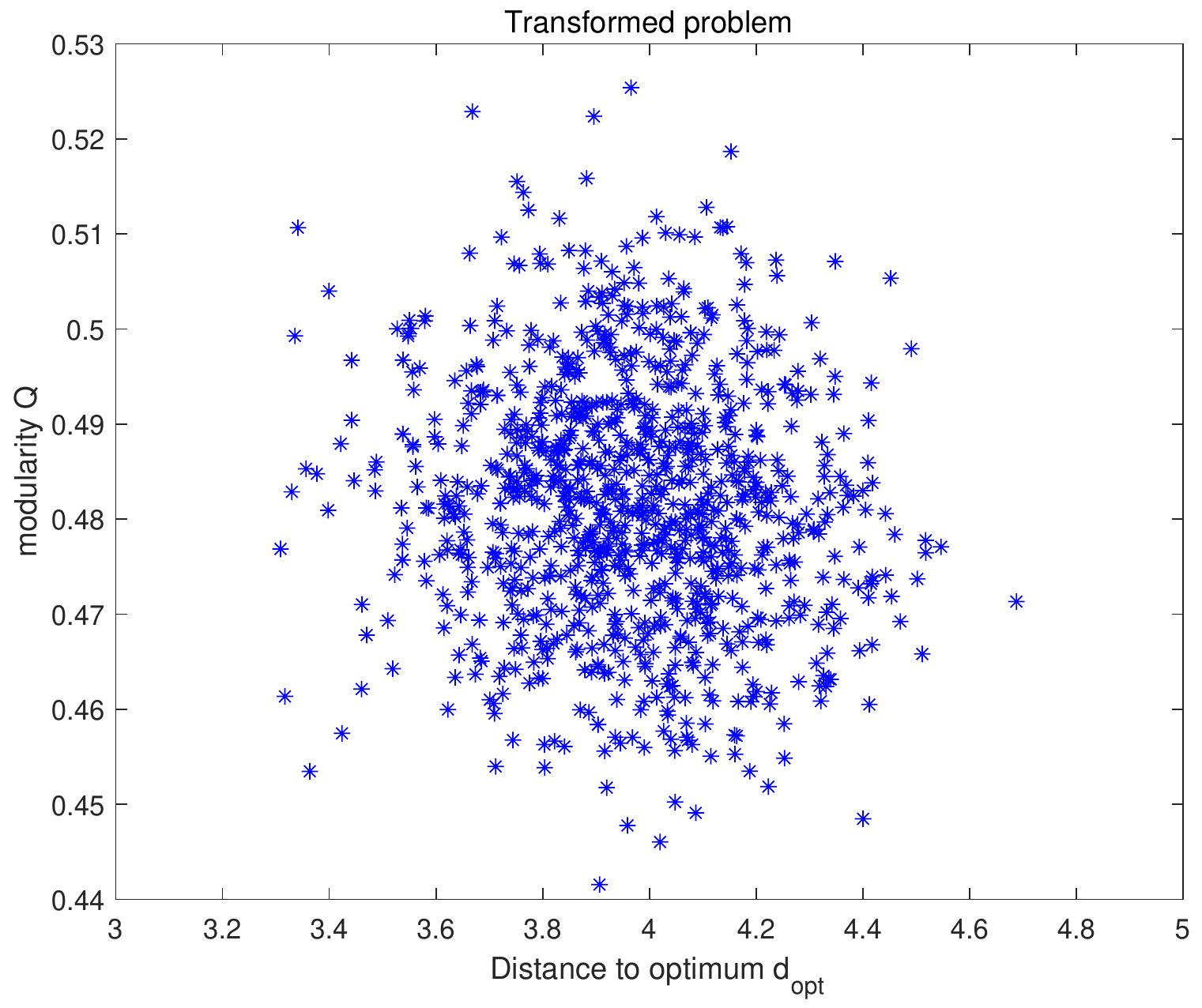}}
\subfigure[]{\includegraphics[width=0.24\linewidth,height=2cm]{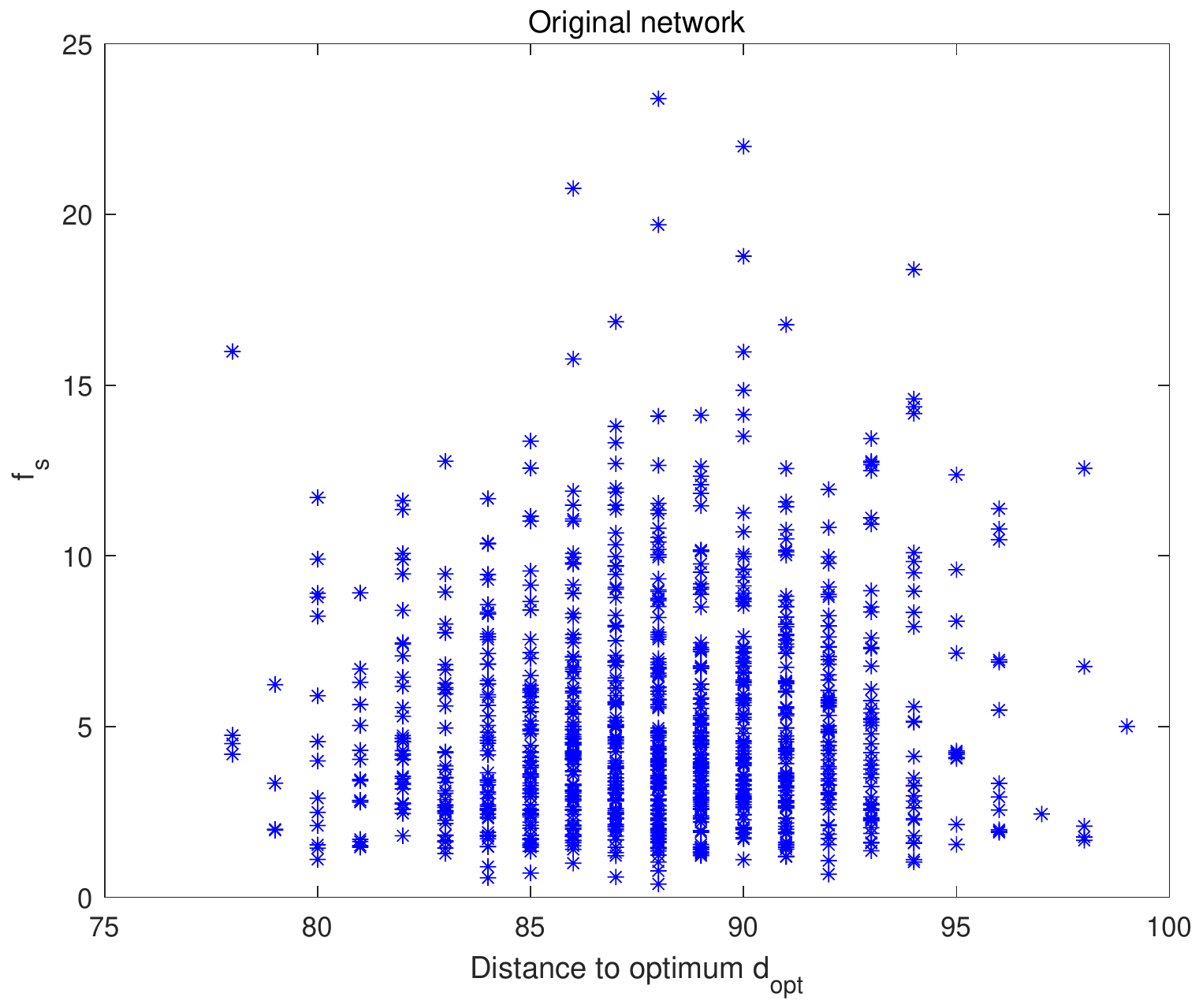}}
\subfigure[]{\includegraphics[width=0.24\linewidth,height=2cm]{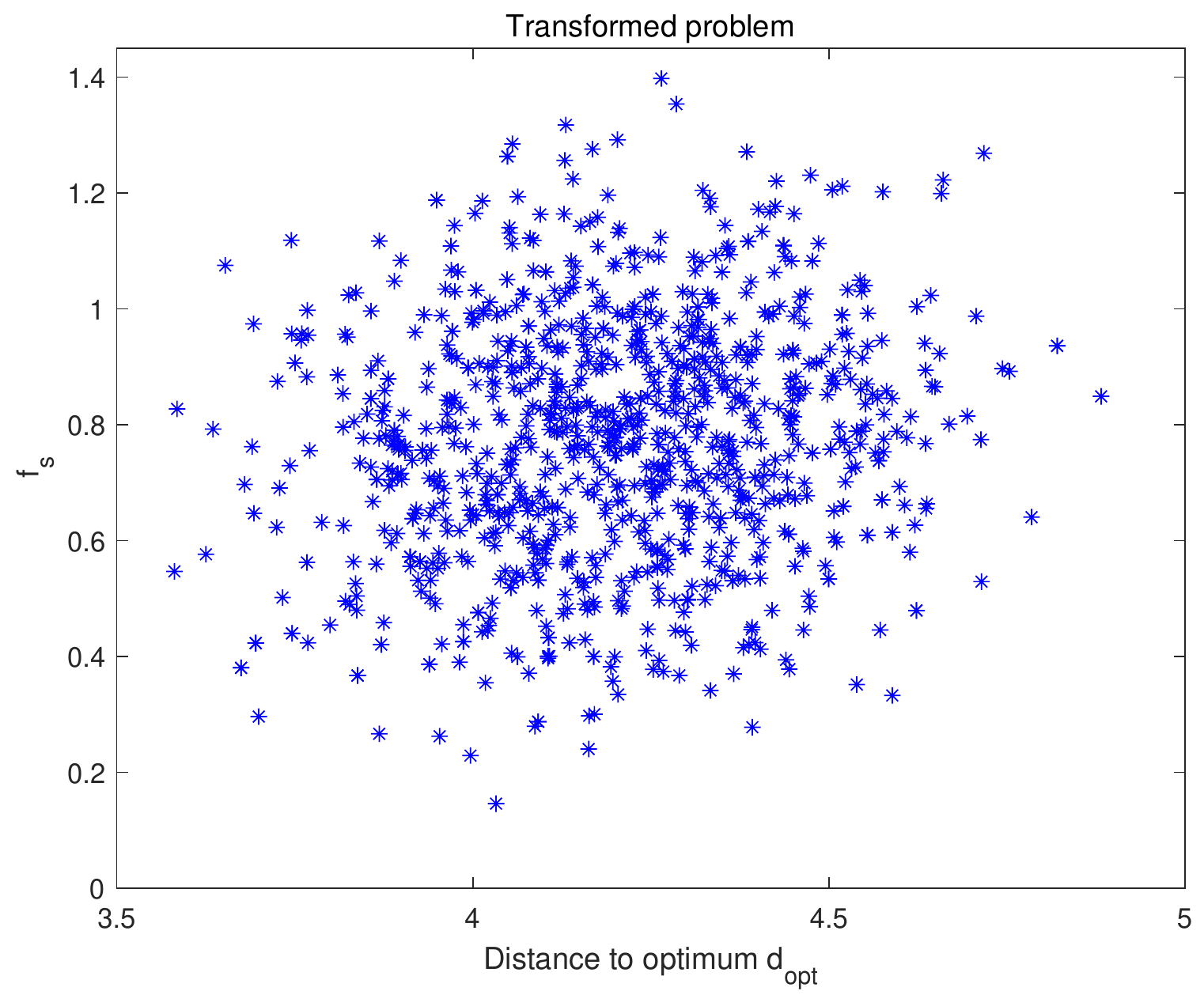}}\\
\caption{The FDC analysis of the original and transformed problems for the Polbooks network. (a) and (b) are for the modularity $Q$, (c) and (d) are for the objective $f_s$. The correlation coefficients between the $x$-axis and $y$-axis values are 0.0475, 0.1326, 0.0666 and 0.2091 for plot (a), (b), (c) and (d), respectively.}
\label{fig5}
\end{figure}

From Tables~\ref{table5} and~\ref{table6}, we find that all the three metrics obtained for the transformed problems are better than those for the original problems on the selected networks. Therefore, we may conclude that the graph network encoding method can smooth the landscape of the community detection problem, which is clearly beneficial to search-based algorithms.

To show the landscape differences, Fig.~\ref{fig5} shows the FDC plots of the original (in subplots (a) and (c)) and transformed problem (in subplots (b) and (d)) on the Polbooks networks. The $x$-axis is the distance to the optimum, while $y$-axis is the objective value. The correlation coefficients between the $x$-axis and $y$-axis for $Q$ are (a) 0.0475 and (b) 0.1326, for $f_s$ are (c) 0.0666 and (d) 0.2091. It is seen that the coefficients obtained for the transformed problems (b and d) are much smaller than those of the original problem (a and c), respectively. This indicates that the landscape of the transformed problem is smoother than that of the original problem.

\section{Related Work}\label{rw}
This section reviews non-EA based methods for community detection in attribute network. These methods can be roughly categorized into two groups, namely distance-based and model-based.

In the distance-based methods, the distance between nodes considering both network structure and attribute homogeneity is key. In~\cite{SA-Cluster}, a graph clustering algorithm, named SA-Cluster, was proposed in which a unified distance metric and a neighborhood random walk distance model was used to estimate the vertex closeness. An improved SA-Cluster, called Inc-Cluster, was proposed to incrementally update the random walk distance~\cite{Inc-Cluster}.

The model-based methods are constructed based on modeling the relationship between network structure and node attributes. One promising method, named BAGC, was proposed in~\cite{BAGC}, in which the community detection for attribute network is modeled under the Bayesian probabilistic framework. Its parameters are estimated by Bayesian inference. A popularity-based conditional link model, called PCL~\cite{PCL}, was proposed to model the node's popularity while the model parameters are estimated by maximum likelihood estimation. In~\cite{TLSC} and~\cite{Jin2019}, two probabilistic generative models, named TLSC and BTLSC respectively, were proposed for topic-related social networks. The generative models are able to distinguish between general and specialized topics. The model parameters are obtained by variational expectation-maximization. In~\cite{circles}, a generative model, named Circles, was proposed using both network structure and node attribute information for ego social networks.

Non-negative matrix factorization (NMF) model was popular for network community detection. In~\cite{SCI}, a NMF model, named SCI, was proposed by taking both the community membership and attribute matrices as decisive variables. Through embedding community structure, attribute network community detection was formulated as a NMF optimization problem in CDE~\cite{CDE}. In~\cite{DCM}, a method named DCM was proposed by alternating between maximizing the community score and inducing a fitting concise description. In~\cite{CYB2019attributed}, a locally weighted $K$-means algorithm, named Adapt-SA, was proposed to learn a fusion weight for each node to balance network structure and node attributes. Fuzzy clustering algorithm was also applied to detect the attribute complex network, such as FCAN~\cite{FCAN}.

Note that all non-EA-based methods mentioned above require the number of communities as {\em a prior}. On the contrary, our method does not need such information which is of more practical use.

\section{Conclusion}\label{h}
In this paper, we proposed a new graph neural network encoding method for complex attribute network community detection problem. Based on the encoding method, the search space of the problem is transformed from discrete to continuous. Our fitness landscape analysis verified that the encoding can smooth the landscape of the original problem for search-based algorithm.

Based on the novel encoding method, combing with two newly developed objectives for single- and multi-attribute similarity respectively and the modularity objective for network structure, we developed a multi-objective evolutionary algorithm, named as CE-MOEA, under the framework of NSGA-II. CE-MOEA was extensively compared against state-of-the-art MOEAs and some well-known non-EA based detection algorithms on a set of real-life networks with different types and with or without true labels. The experimental results clearly showed that CE-MOEA performed significantly better than MOEA-SA, MOGA-@Net and those non-EA algorithms in general. Particularly, since MOEA-SA, MOGA-@Net and CE-MOEA were built upon NSGA-II, the superior performance of CE-MOEA implied that the developed graph neural network encoding is beneficial for the optimization.

In the future, we intend to 1) apply the graph neural network encoding method for overlapping complex attribute network community detection; 2) develop specific neural network encoding for other discrete optimization problems such as traveling salesman problem, and others.


\section*{Acknowledgement}

The authors would like to thank Prof J. Liu and Prof C. Pizzuti for providing the codes of their MOEAs.



\bibliographystyle{IEEEtran}
\bibliography{IEEEbib}
\end{document}